\documentclass[letterpaper, 10 pt, conference]{ieeeconf}  

\IEEEoverridecommandlockouts                              

\usepackage{cite}
\usepackage{amsmath,amssymb,amsfonts}
\usepackage{algorithmic}
\usepackage{graphicx}
\usepackage{textcomp}
\usepackage{multirow}
\usepackage{xcolor}
\usepackage{leftidx}
\usepackage{color,soul}
\usepackage{caption}
\usepackage{subcaption}
\usepackage{makecell}
\usepackage[export]{adjustbox}
\usepackage{hyperref}
\usepackage{booktabs}
\usepackage[nolist]{acronym}
\usepackage{tikz}
\usepackage{tabularx}
\usepackage{makecell}
\usepackage{lipsum}
\usepackage{gensymb}
\usepackage{mathtools}
\usepackage{duckuments}
\usepackage{siunitx}
\usepackage{verbatim}

\let\labelindent\relax
\usepackage[inline]{enumitem}

\newacro{slam}[SLAM]{Simultaneous Localization and Mapping}
\newacro{uav}[UAV]{Unmanned Aerial Vehicle}
\acrodefplural{uav}[UAVs]{unmanned aerial vehicles}
\newacro{gns}[GNS]{Global Navigation Satellite}
\newacro{gnss}[GNSS]{Global Navigation Satellite System}
\acrodefplural{gnss}[GNSS]{Global Navigation Satellite Systems}
\newacro{mcl}[MCL]{Monte-Carlo localization}
\newacro{imu}[IMU]{Inertial Measurement Unit}
\newacro{dof}[DOF]{degree-of-freedom}
\acrodefplural{dof}[DOFs]{degrees-of-freedom}
\newacro{ransac}[RANSAC]{random sample consensus}
\newacro{map}[MAP]{maximum a posteriori}
\newacro{mle}[MLE]{maximum likelihood estimation}
\newacro{rms}[RMS]{root-mean-square}
\newacro{dem}[DEM]{digital elevation model}
\newacro{vio}[VIO]{visual-inertial odometry}
\newacro{cnn}[CNN]{convolutional neural network}
\newacro{pdf}[pdf]{probability density function}
\newacro{ahrs}[AHRS]{attitude and heading reference system}
\newacro{lidar}[LIDAR]{light detection and ranging}
\newacro{relu}[ReLU]{rectified linear unit}
\newacro{rtk}[RTK]{real-time kinematic}
\newacro{gps}[GPS]{global positioning system}
\newacro{fcn}[FCN]{fully-connected network}
\newacro{brm}[BRM]{building ratio map}

\newcommand{\figref}[1]{\hyperref[#1]{Fig.~\ref*{#1}}}
\newcommand{\tabref}[1]{\hyperref[#1]{Tab.~\ref*{#1}}}
\newcommand{\secref}[1]{\hyperref[#1]{Sec.~\ref*{#1}}}
\newcommand{\algoref}[1]{\hyperref[#1]{Alg.~\ref*{#1}}}

\newcommand{\cond}[2]{p(#1\vert#2)}

\newcommand{\boldx}{\mathbf{X}}
\newcommand{\weightmatrix}{\mathbf{W}}
\newcommand{\descriptorvector}{\mathbf{w}}
\newcommand{\bigY}[1]{\mathbf{Y}_{#1}}
\newcommand{\continuousX}{\mathbf{\mathcal{X}}}
\newcommand{\kernel}{\mathbf{\kappa}}

\newcommand{\squarecoordinateframe}[1]{$\{C_{s,#1}\}$}
\newcommand{\localcoordinateframe}[1]{$\{C_{l,#1}\}$}

\def\xy{$(x, y)$}
\def\xycoords{\xy-coordinates}

\def\ground{ground-truth}
\def\Ground{Ground-truth}

\def\sota{state-of-the-art}
\def\ie{\textit{i.e.},}
\def\eg{\textit{e.g.},}

\def\etal{\textit{et al.}}

\def\km2{km$^2$}

\def\precomputedmap{\mathbf{M}}

\def\maparea{100 km$^2$}
\def\translationbetweenupdates{50}
\def\localcoordtosquarecoord{T_{l,k}^{s,k}}

\def\squaresizem{100m by 100m}
\def\squaresizepx{100px by 100px}
\def\likelihoodconversiontableheader{Likelihood conversion}
\def\modeltypetableheader{Model type}
\def\translerrorafterconvergence{12.6--18.7 m}
\def\numberofstepstoconvergence{23.2--44.4}

\def\kauhavamapdimensions{1.62 km by 3.82 km}

\def\figvspace{\vspace{-1.3em}}

\newcolumntype{L}[1]{>{\raggedright\arraybackslash}p{#1}}
\newcolumntype{C}[1]{>{\centering\arraybackslash}p{#1}}
\newcolumntype{R}[1]{>{\raggedleft\arraybackslash}p{#1}}

\def\methodname{LSVL}

\title{\LARGE \bf
\methodname{}: Large-scale season-invariant visual localization for UAVs
}

\author{Jouko Kinnari$^{1}$, Riccardo Renzulli$^{2}$, Francesco Verdoja$^{3}$ and Ville Kyrki$^{3}$
\thanks{This work was supported by Business Finland project Multico (6575/31/2019) and Saab Finland Oy.}
\thanks{$^{1}$J. Kinnari is with Saab Finland Oy,
Salomonkatu 17B, 00100 Helsinki, Finland
{\tt\small \{firstname.lastname\}@saabgroup.com}}%
\thanks{$^{2}$R. Renzulli is with Department of Computer Science, University of Turin, Italy {\tt\small \{firstname.lastname\}@unito.it}}
\thanks{$^{3}$F. Verdoja and V. Kyrki are with School of Electrical Engineering, Aalto University, Finland {\tt\small \{firstname.lastname\}@aalto.fi}}%
}

\begin{document}

\maketitle


\begin{abstract}
Localization of autonomous \acp{uav} relies heavily on \acp{gnss}, which are susceptible to interference. Especially in security applications, robust localization algorithms independent of \ac{gnss} are needed to provide dependable operations of autonomous \acp{uav} also in interfered conditions. Typical non-\ac{gnss} visual localization approaches rely on known starting pose, work only on a small-sized map, or require known flight paths before a mission starts. We consider the problem of localization with no information on initial pose or planned flight path. We propose a solution for global visual localization on a map at scale up to \maparea{}, based on matching orthoprojected UAV images to satellite imagery using learned season-invariant descriptors. We show that the method is able to determine heading, latitude and longitude of the \ac{uav} at \translerrorafterconvergence{} lateral translation error in as few as \numberofstepstoconvergence{} updates from an uninformed initialization, also in situations of significant seasonal appearance difference (winter-summer) between the \ac{uav} image and the map. We evaluate the characteristics of multiple neural network architectures for generating the descriptors, and likelihood estimation methods that are able to provide fast convergence and low localization error. We also evaluate the operation of the algorithm using real \ac{uav} data and evaluate running time on a real-time embedded platform. We believe this is the first work that is able to recover the pose of an \ac{uav} at this scale and rate of convergence, while allowing significant seasonal difference between camera observations and map.
\end{abstract}

\acresetall

\section{Introduction}
The ability of a \ac{uav} to robustly estimate its position is one of the basic requirements of autonomous flight. In missions in which the \ac{uav} needs to collaborate with other agents operating in the same environment, information of position in a shared, global frame of reference is needed.

There are several possible ways to estimate position. A localization solution can rely on infrastructure built for this purpose. In outdoor \ac{uav} operations, by far the most common is to rely on \acp{gnss}. In ideal conditions, \acp{gnss} receivers provide measurements of position. However, \acp{gnss}, as well as other radio beacon systems, are susceptible to blockages in radio signal path, inaccuracies due to multipath propagation, and spoofing and jamming attacks \cite{7445815, RR-2970-DHS}, which may be used for denying operation of \acp{uav} in an area.

A sensor system commonly carried by \acp{uav} is the combination of a camera and an \ac{imu}. Several works have focused on using this sensor combination for resolving the position of the \ac{uav} by tracking the difference to a known starting position by methods of \ac{vio} \cite{Scaramuzza2020}. As \ac{vio} implementations integrate noisy signals, they suffer from drift over long flights \cite{8460664}. \ac{slam} methods \cite{7747236} can help in partially compensating for this drift by detecting loop closures, but only if the mission contains reentry to a previously visited area. In addition, \ac{vio} and \ac{slam} approaches are not robust to random failures in position tracking; temporary failures may lead to loss of position information in the global frame with only a small chance of recovery.

\begin{figure}
     \centering
     \includegraphics[width=\linewidth]{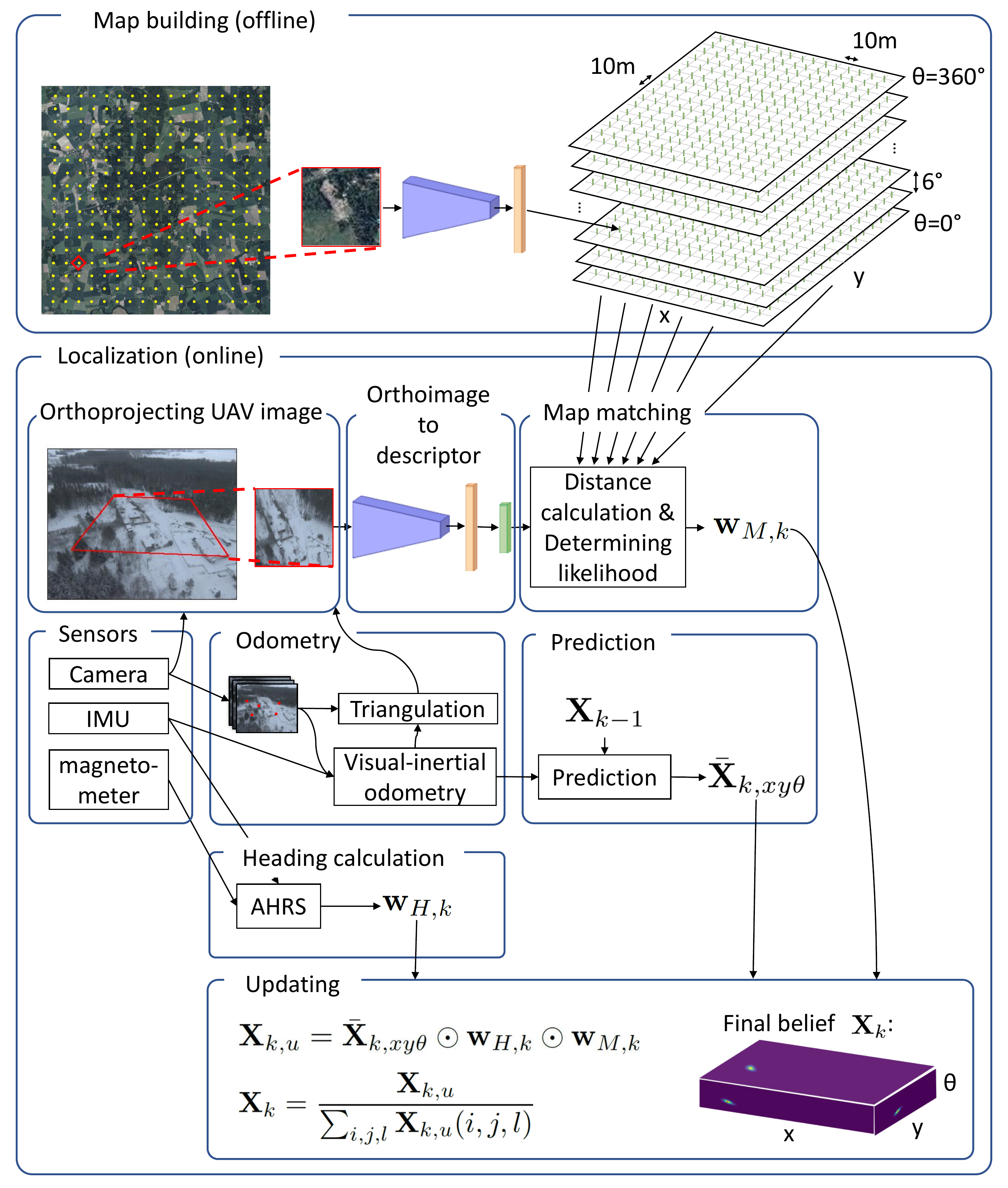}
     \caption{Block diagram of proposed localization solution.}
     \label{fig:first_page_image}
\end{figure}

If a map is available, matching camera images acquired by a \ac{uav} to the map allows compensating for drift induced by odometry methods and provides an estimate of position with respect to the map. This approach is called visual localization \cite{10.1145/3281548.3281556, COUTURIER2021103666}. Using a georeferenced map allows, in principle, finding the position of a \ac{uav} with respect to the map, even in the case of no prior information on position at start of mission or after random failures in positioning.

There are, however, multiple challenges in visual localization. Appearance difference between the \ac{uav} image and the map may be significant due to changes in season --- an image acquired by a \ac{uav} in winter looks very different from a satellite image acquired in summer. Moreover, the environment in which the \ac{uav} is operating may be naturally ambiguous, \eg{} when flying over vast areas of forests. In order for the localization solution to support recovery from random localization failures, the size of the map must correspond to the operating area of the \ac{uav}. This means that the localization solution must be computationally efficient enough to run on an onboard computer even on large maps, and it must tolerate the natural ambiguity.

In this work we propose \methodname{}, which addresses all of the above mentioned challenges of robust visual localization at large scale. The main contributions of this work are:
\begin{enumerate}
    \item We propose a \ac{uav} image to map matching solution based on descriptors learned in a manner that provides invariance to seasonal appearance difference. We train the descriptor networks using only satellite images, without requiring any labeled data such as semantic classification of the terrain.
    \item We present the problem of \ac{uav} visual localization as state estimation utilizing a point mass filter, and integrate it to our map matching approach. This combination allows running our algorithm in real time, with constant time and memory consumption onboard a \ac{uav} at sufficiently large scale for \ac{uav} missions.
    \item We explore different architecture choices for image description, vector dimensionality, and likelihood computation, and evaluate their impact on probability to convergence, time to convergence and positioning error after convergence, on real data.
    \item We show that our approach is able to tolerate natural ambiguities and resolve location of a \ac{uav}, starting from an uncertainty corresponding to an area of \maparea{}, converging to an average translation error of \translerrorafterconvergence{} after no more than \numberofstepstoconvergence{} updates with \ac{uav} camera observations when flying over ambiguous terrains under significant seasonal appearance difference between \ac{uav} image and map. Our formulation does not require a digital elevation model and can operate using 2D maps.
    \item We present a simple method allowing assessing probability of convergence to true pose, enabling self-diagnostics of the localization solution, which is a key component in the robust global localization problem.
    \item We compare our solution to two \sota{} approaches for \ac{uav} localization.
    \item We demonstrate the operation of our localization solution in real time onboard a commercial \ac{uav}.
\end{enumerate}
We believe this is the first work in the visual localization area that is able to find true position of the \ac{uav}, starting from a scale of uncertainty of \maparea{}, over ambiguous terrains, under significant seasonal appearance change. A block diagram of our algorithm is shown in \autoref{fig:first_page_image}.

The paper is structured as follows. \secref{sec:related_work} goes over related work. \secref{sec:preliminaries} defines preliminaries of the localization problem. We introduce our method in \secref{sec:method} and detail a vital part of the solution, map matching, in \secref{sec:map_matching}. \secref{sec:experiments} describes our localization experiments and we conclude the paper with discussion and conclusions in \secref{sec:discussion} and \secref{sec:conclusions}, respectively.

\section{Related work}
\label{sec:related_work}
Visual localization of \acp{uav} is a topic which has attracted interest especially over past few years. The \sota{} is covered by relatively recent surveys \cite{10.1145/3281548.3281556, COUTURIER2021103666}. Within the area of visual localization, most works are limited by one or more of the following assumptions: accurate initialization is required \cite{6942633,9196606,9357892,8793558}, the size of the operating area is limited \cite{7418753,9636705}, the movements of the \ac{uav} are constrained to specific paths \cite{9357892}, the operation takes place in conditions that are very close to map in terms of appearance \cite{6942633,7418753}, the map resolution and detail requirement is significant,  such as requiring a topography map \cite{9636705}, or a very high flying altitude is required for successful georeferencing \cite{9170807}. The following overview of related work focuses on the works that---similarly to \methodname{}---do not require of knowledge of initial pose and that use an easily obtainable planar 2D map for localization.

A common choice in \ac{uav} localization is to detect semantic features such as roads and intersections \cite{Dumble2015AirborneVN,volkova2018more,9184263,9170807} or buildings \cite{9341682}. Choi \etal \cite{9341682} proposed a method where the \ac{uav} image is semantically segmented to find buildings in the camera view. Based on detected buildings, a rotation invariant descriptor, \ac{brm}, is computed from the proportion of building pixels visible in camera view, and a precomputed map with similar descriptors is used for localization. The authors demonstrated convergence of position estimate on a 6.17 $km^2$ map after 27 updates with 25 meters of translation between each update, with 12.01 m \ac{rms} error after convergence. The demonstration flight takes place over a residential area. The main drawback of this approach is the requirement that features of a specific semantic class (\ie{} buildings) are required for successful localization, which may be unavailable when flying over natural environments.

Mantelli \etal{} \cite{MANTELLI2019304} demonstrate localizing a \ac{uav} on a map of size 1.34 \km2{} using a particle filter, where particle likelihood is determined based on a handcrafted descriptor called abBRIEF. The authors initialize the particle filter with 50 000 particles and show robustness of their localization solution in comparison to BRIEF \cite{calonder2010brief} descriptors on trajectory lengths up to 2.4 km, showing convergence in less than 50 m to an average translation error of 17.78 m. The descriptor is developed such that it is tolerant to illumination changes and allows fast computation of particle likelihood over a large number of pose hypotheses.

We evaluate against \ac{brm} and abBRIEF and show superior performance.

We build on the idea of compressing \ac{uav} camera observation into a compact embedding space, to allow fast testing of pose hypotheses on a large map. This has been proposed recently by \eg{} Bianchi \etal{} \cite{9357892} who use a bottleneck autoencoder approach to compress visual observations to descriptor vectors of dimension 1000. Also Samano \etal{} \cite{9562005} and Couturier \etal{} \cite{10.1117/12.2585986} train Resnet models \cite{7780459} to project map tiles and \ac{uav} images to a low-dimensional (16D) embedding space. Only Samano \etal \cite{9562005} allow movements of the \ac{uav} outside precomputed paths by precomputing a grid of embedding vectors, on which hypothesis testing is performed by interpolating a vector from this grid by each pose hypothesis and measuring distance of observed image descriptor to interpolated descriptor vector in embedding space. Authors of \cite{9562005} demonstrate with simulated flight experiment the convergence of pose estimate to less than 95 m translation error in 78.2\% of simulated flights by 200 updates. We take a similar approach but forgo interpolation in embedding space, to avoid unjustified implicit assumption of smoothness of embedding space, and explore other learned descriptor architectures. Compared to \cite{9562005}, we show significantly faster convergence, tolerance to seasonal variation and we demonstrate performance with real experiments.

\section{Preliminaries}
\label{sec:preliminaries}

We want to resolve the pose of an \ac{uav} using measurements available at the \ac{uav} during flight, with no dependency on localization or communication infrastructures. Formally, we define the pose of the \ac{uav} pose in a common, global reference frame as the state
\begin{equation}
\continuousX_k = \begin{bmatrix}x_k & y_k & \theta_k\end{bmatrix}^T
\end{equation}
where $x$, $y$, and $\theta$ represent longitude, latitude, and heading, respectively, in a Cartesian coordinate system and $k$ is index for time. We consider localization of the \ac{uav} in a limited region in longitude and latitude: we assume $x_k \in [x_{min}, x_{max}]$, $y_k \in [y_{min}, y_{max}]$, $\theta_k \in [0, 2\pi)$. The localization problem is to compute the marginal posterior distribution $\cond{\continuousX_k}{\bigY{1:k},\mathcal{M}}$ of state $\continuousX_k$, given history of measurements $\bigY{1:k}$ and map $\mathcal{M}$.

In this work, we concentrate on the \emph{wake-up robot problem}, \ie{} at the start of a flight, we assume an uniform prior distribution $p(\continuousX_0)$ over all values of $x_k$ and $y_k$, across all values of heading. This represents a situation where the only initial information about the \ac{uav} position is that it is located within the area of a map defined over a rectangular area in latitude and longitude. Our formulation allows inclusion of more informed initialization.

We take the typical approach of considering localization on preacquired map as a Bayesian filtering problem (see \eg{} \cite{thrun2002probabilistic}). This amounts to maintaining a representation of belief of current state and updating that representation when new measurements are available. At each time step $k$, we obtain three types of measurements from the onboard sensors of the \ac{uav}: an odometry measurement $\mathbf{u}_k$, an heading measurement $v_k$, and an image from the \ac{uav} camera $\mathcal{I}_k$ to be used for map matching. Given the measurment $\bigY{k} = \begin{Bmatrix}
\mathbf{u}_k, v_k, \mathcal{I}_k
\end{Bmatrix}$, at each sampling time $k$, we first perform a prediction step based on the odometry measurement:
\begin{multline}
\label{eq:prediction}
    \cond{\continuousX_k}{\mathbf{u}_k,\bigY{1:k-1}} = \\
    \int \cond{\continuousX_k}{\mathbf{u}_k,\continuousX_{k-1}} \cond{\continuousX_{k-1}}{\bigY{1:k-1}} d\continuousX_{k-1}
\end{multline}
We then use the heading and map matching measurements as our observation model:
\begin{equation}
\label{eq:state_estimate_continuous}
\cond{\continuousX_k}{\bigY{1:k}} = \frac{1}{\eta_k} \cond{v_k, \mathcal{I}_k}{\continuousX_k,\mathcal{M}} \cond{\continuousX_k}{\mathbf{u}_k,\bigY{1:k-1}}
\end{equation}
where $\eta_k$ is a normalizing constant. The heading and map matching measurements are considered independent, and likelihood of heading measurement is not conditional on map:
\begin{equation}
\cond{v_k, \mathcal{I}_k}{\continuousX_k,\mathcal{M}} = \cond{v_k}{\continuousX_k}\cond{\mathcal{I}_k}{\continuousX_k,\mathcal{M}}
\end{equation}

\section{Method}
\label{sec:method}

In order to find the pose on a large map in presence of matching ambiguities presented in the introduction, a method is needed for utilizing a sequence of as many \ac{uav} images as needed in order to converge to a single, correct pose estimate. We propose a recursive \emph{localization method} consisting of the following components, illustrated in \figref{fig:first_page_image}. We use an \emph{odometry measurement} for predicting belief of state $\bar{\boldx}_{k,xy\theta}$ at time $k$ based on state $\boldx_{k-1}$ at previous time instant $k-1$. We use a \emph{map matching measurement} for computing likelihood of pose hypotheses $\weightmatrix_{M,k}$ based on a single \ac{uav} image and, optionally, a \emph{heading measurement} for computing the likelihood of pose hypotheses $\weightmatrix_{H,k}$ based on a compass heading measurement. The localization method updates belief of state using all the measurements, providing $\boldx_{k}$, belief of state at time $k$.

In this section we describe each of these components in detail, putting particular attention in describing how the proposed solutions target the complexity that arises from both dealing with a large map and with considerable appearance change, central limitations of current methods that we address in this work. \secref{sec:measurements} lists all measurements we use, and \secref{sec:localization_method} specifies our localization method.

\subsection{Measurements}
\label{sec:measurements}
\subsubsection{Odometry measurement}
\label{sec:odometry_measurement}

We assume that the \ac{uav} is running a \ac{vio} algorithm which, at time $k$, provides $\mathbf{u}_k = \begin{bmatrix}u_{k,x} & u_{k,y} & u_{k,\theta} & u_{k,o}\end{bmatrix}^T$, a measurement of translation, rotation and distance traveled since time $k-1$. $u_{k,x}$ and $u_{k,y}$ are translation since time instant $k-1$ in $x$ and $y$ coordinates, respectively, with respect to pose at time $k-1$. Similarly, $u_{k,\theta}$ is rotation around vertical axis. $u_{k,o}$ is the integral of distance traveled since instant $k-1$ according to odometry. The measurements are visualized in \figref{fig:odometry_measurements}.

We approximate the odometry pose uncertainty with a multivariate normal distribution. We further assume the covariance of pose is isotropic in $x$, $y$ directions and that rotation noise is independent from translation noise. More formally, we assume $\mathbf{u}_k$ has posterior density $p(\continuousX_k | \mathbf{u}_k, \continuousX_{k-1})$ over possible successor states $\continuousX_k$, given the previous state $\continuousX_{k-1}$ and the odometry measurement $\mathbf{u}_k$ and we approximate the posterior density with a multivariate normal distribution with covariance $\mathbf{\Sigma}_u$:
\begin{equation}
\label{eq:odometry_posterior_density}
    p(\continuousX_k | \mathbf{u}_k, \continuousX_{k-1}) = \mathcal{N}(\continuousX_{k-1}+\begin{bmatrix}u_{k,x} \\ u_{k,y} \\ u_{k,\theta}\end{bmatrix},\mathbf{\Sigma}_{u}(u_{k,o}))
\end{equation}
where $\mathbf{\Sigma}_u(t)$ is a diagonal matrix:
\begin{equation}
    \mathbf{\Sigma}_u(t) = diag(\sigma_{u,xy}(t)^2, \sigma_{u,xy}(t)^2, \sigma_{u,\theta}(t)^2).   
\end{equation}
While the isotropic noise for translation and independent noise in heading are simplified approximations of the true distribution \cite{long2013banana}, we consider this a sufficient upper bound approximation to the odometry noise. This approximation enables a computationally fast method of prediction as elaborated in \secref{sec:using_odometry_for_prediction}.

In addition, we assume the movements of the \ac{uav} produce sufficient \ac{imu} excitation to make scale observable or that the \ac{uav} is equipped with additional sensors with which scale can be resolved.

\subsubsection{Map matching measurement}

The purpose of map matching is to provide a means for verifying or disputing pose hypotheses, given camera image $\mathcal{I}_k$ and map $\mathcal{M}$. A method for computing the likelihood $\cond{\mathcal{I}_k}{\continuousX_k,\mathcal{M}}$ is thus needed. Our work focuses especially on finding means for computing the likelihood that work well over ambiguous terrains under significant seasonal appearance difference, in a computationally efficient way. Since the method used for map matching and likelihood computation based on a \ac{uav} image is a considerable part of our contribution, a separate section (\secref{sec:map_matching}) has been given to the detailed description of this component.

\subsubsection{Heading measurement}
\label{sec:heading_measurement}

We assume the \ac{uav} is equipped with an \ac{ahrs}, relying on fusion of a compass and an \ac{imu}, which provides a measurement of heading, $\theta$, with respect to map East, corrupted by Gaussian noise $n_\theta \sim \mathcal{N}(0,\sigma_{v}^2)$:
\begin{equation}
\label{eq:heading_measurement}
    v_k = v(t_k) = \theta(t_k) + n_{\theta}
\end{equation}

\subsection{Localization method}
\label{sec:localization_method}
We have to consider how to formulate the pose estimation problem in a computationally feasible way. In choosing the estimation approach, we need to consider characteristics of the problem. Specifically, as we start from a very uninformed state and expect natural ambiguities in the environment we are operating in, it is expected that before converging to the correct pose, our estimator has to be able to track a large number of multiple possible hypotheses.

\subsubsection{Choosing representation for state}

Typical solutions used in multimodal location estimation problems include particle filtering and point mass filtering \cite{thrun2002probabilistic}. The use of a particle filter in \ac{uav} localization is a common choice \cite{9562005, 7418753, MANTELLI2019304}. However, in cases of very uninformed initialization, a risk exists that a particle is initially not placed in vicinity of the true pose, leaving the probability of converging on the correct pose over time to chance. Furthermore, during a flight, it is possible that the close proximity of true state is left without sufficient particle density in cases such as when flying for long periods of time over ambiguous areas, or in case of large but local map inconsistencies (e.g. forest clearcutting having taken place between map acquisition and flight), leading to divergence in a poorly predictable way. Since the particle filter is stochastic, different instantiations of the filter using the same data may provide different results and, depending on selected resampling scheme, computation time may vary.

Instead of representing belief through the use of particles, we choose to use a point mass filter and compute the belief on a discrete grid. This selection ensures coverage of full state space throughout the mission, at the resolution specified by our choice of grid, and offers deterministic performance and constant computational time and memory requirement. Early adaptations of point mass filtering-based approaches on 2D robot localization include the work by Burgard \etal{} \cite{10.5555/1864519.1864520} and in terrain navigation of flying platforms by Bergman \cite{bergman1999recursive}.

\subsubsection{Definition of state using point mass filter}
We approximate belief $p(\continuousX)$ on the continuous state space of $\continuousX$ by decomposing it into a grid of regions of equal size with resolution $r_x = r_y = r_{xy}$ in translation and $r_{\theta}$ in heading. Each region is a voxel $s(i,j,l) = \begin{bmatrix}s_x(i),s_y(j),s_\theta(l)\end{bmatrix}$ in the state space with bounds
\begin{equation}
\begin{aligned}
    s_{xv}(i) &\leq s_x(i) < s_{xu}(i) \\
    s_{yv}(j) &\leq s_y(j) < s_{yu}(j) \\
    s_{\theta v}(l) &\leq s_\theta(l) < s_{\theta u}(l)
\end{aligned}
\end{equation}
with lower bound $s_{bv}(i) = b_{min} + i \times r_{b}$ and upper bound $s_{bu}(i) = s_{bv}(i) + r_{b}$ for each axis $b \in \{x,y,\theta\}$ and $i, j, l \in \mathbb{N}$ such that the whole state space is covered.

We approximate belief over state at time instant $k$ as a piecewise constant probability matrix $\boldx_k \approx p(\continuousX_k)$
where each element of matrix $\boldx_k[i,j,l]$ assigns a probability to each voxel $s(i,j,l)$ in state space.

\subsubsection{Using odometry measurement for prediction}
\label{sec:using_odometry_for_prediction}
\begin{figure}
\centering
     \begin{subfigure}[t]{0.8\linewidth}
         \centering
         \includegraphics[width=\linewidth]{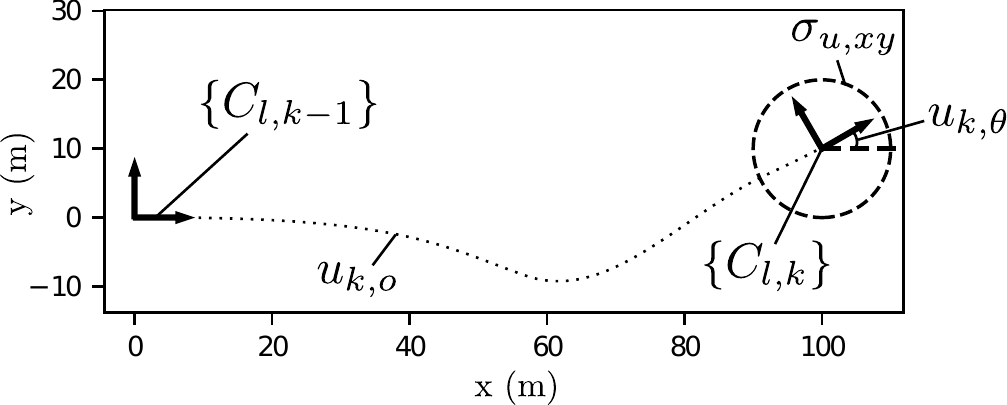}
         \caption{Odometry measurements are stated with respect to frame at previous update, \localcoordinateframe{k-1}. $u_{k,o}$ states distance traveled since previous update, according to odometry.}
         \label{fig:odometry_measurements}
    \end{subfigure}
    
    \begin{subfigure}[t]{\linewidth}
        \centering
        \includegraphics[width=0.7\linewidth]{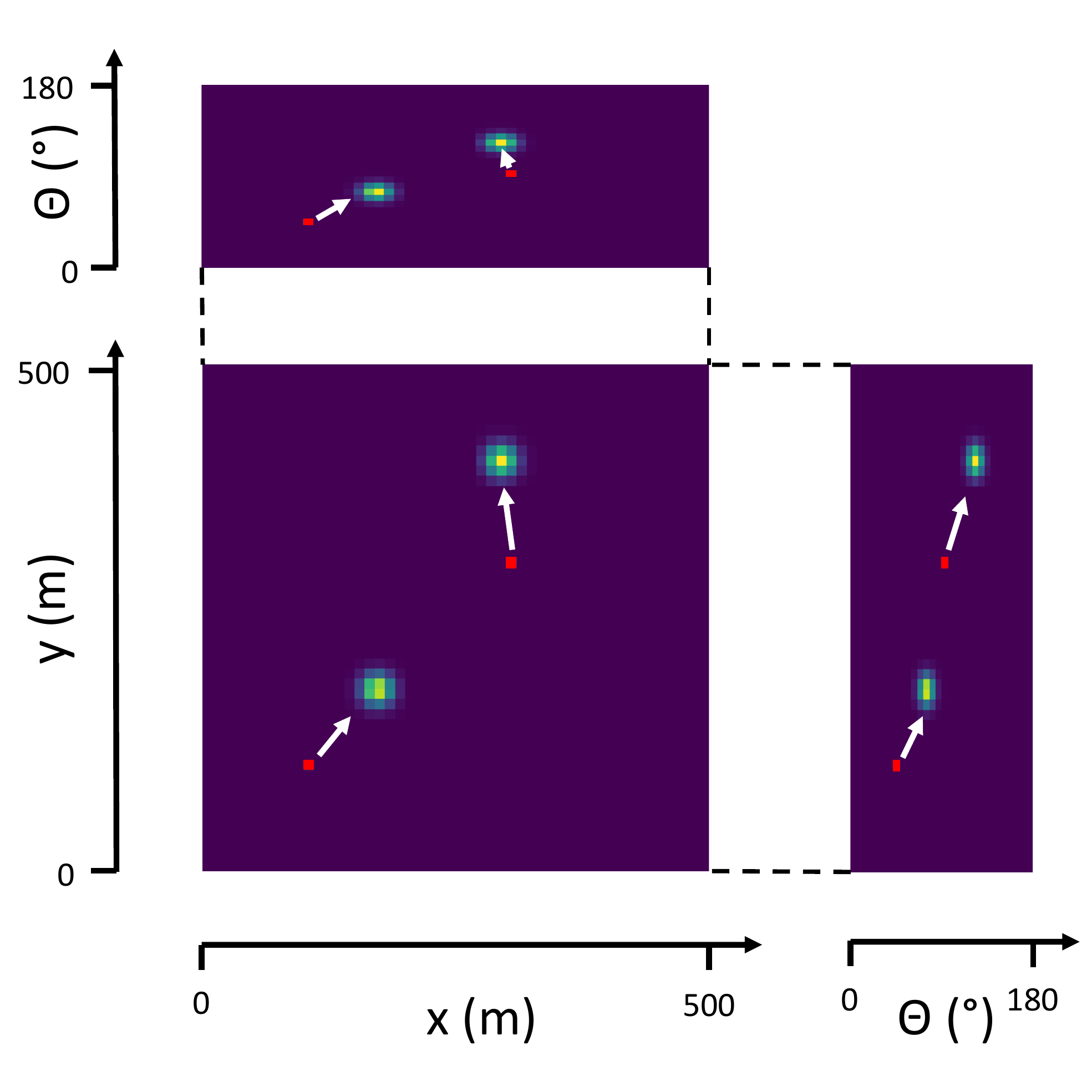}
        \caption{Part of $\boldx$ after marginalizing different axes individually. Two voxels at $(100,100,45\degree)$ and $(300,300,90\degree)$ representing belief before prediction are shown in red. Voxels highlighted in green and yellow represent belief after prediction. White arrows show how prediction using these odometry measurements with the assumed noise shifts and smooths $\boldx$ according to the measurement, and that shifts are performed in direction determined by odometry measurement and voxel's $\theta$ value.}
        \label{fig:x_marginalized_odometry}
    \end{subfigure}
    
    \caption{Example of prediction based on odometry. In this example, $u_{k,x}=100$ m, $u_{k,y}=10$ m, $u_{k,\theta}=30\degree$, $\sigma_{u,xy} = 10$ m, $\sigma_{u,\theta} = 5\degree$. \figref{fig:odometry_measurements} visualizes odometry measurements between times $k-1$ and $k$. \figref{fig:x_marginalized_odometry} visualizes how odometry measurements are used in prediction.}
    \label{fig:odometry_updating_example}
\end{figure}

To approximate the prediction step \eqref{eq:prediction} with our state representation, using the \ac{vio} measurement model presented in \secref{sec:odometry_measurement}, we formulate a method in which the probability mass contained in a voxel in $\boldx_{k-1}$ is projected to other voxels in the belief grid at time $k$ as dictated by the odometry measurement. We repeat this operation for each voxel in the belief grid. An example visualization is shown in \figref{fig:odometry_updating_example}. Isotropic \xy{} odometry noise and independent $\theta$ odometry noise enable us to run the prediction computationally efficiently as three consecutive 1D convolutions.

We compute offsets $o_x(\alpha)$, $o_y(\alpha)$, $o_\theta$ and kernel vectors $\kernel_\theta$, $\kernel_x(\alpha)$ and $\kernel_y(\alpha)$ such that for a given initial heading $\alpha$ in the global frame, the kernels span a region of at least four standard deviations around mean. We then perform prediction by running 1D convolutions in sequence: \begin{subequations}
    \begin{equation}
        \label{eq:x_convolution}
        \bar{\boldx}_{k,x}[i,j,l] = \sum_{h=1}^{q_x} \boldx_{k-1}[i - o_x(\bar{s}_\theta(l)) - h,j,l] \kernel_x(\bar{s}_\theta(l))[h]
    \end{equation}
    \begin{equation}
        \label{eq:y_convolution}
        \bar{\boldx}_{k,xy}[i,j,l] = \sum_{h=1}^{q_y} \bar{\boldx}_{k,x}[i,j - o_y(\bar{s}_\theta(l)) - h,l] \kernel_y(\bar{s}_\theta(l))[h]
    \end{equation}
    \begin{equation}
        \label{eq:theta_convolution}
        \bar{\boldx}_{k,xy\theta}[i,j,l] = \sum_{h=1}^{q_\theta} \bar{\boldx}_{k,xy}[i,j,l - o_\theta - h] \kernel_\theta[h]
    \end{equation}
\end{subequations}
Here, $\bar{s}_\theta(l)$ returns the centerpoint of the angle corresponding with index $l$ in $\boldx$. Note that the offsets and convolution kernel vectors in $x$ and $y$ direction depend on the value of $\alpha$, \ie{} probability mass is shifted in the direction defined by the heading value, using the nominal value of that cell. At the edges of state space, \eqref{eq:x_convolution} and \eqref{eq:y_convolution} are filled with zero for values outside state space and \eqref{eq:theta_convolution} is wrapped around to the opposite edge.

The end result of this step is $\bar{\boldx}_{k,xy\theta}$, which states how $\boldx_{k-1}$ shifted and spread according to odometry measurement and odometry noise from time $k-1$ to $k$. A clarifying visualization can be found in \figref{fig:odometry_updating_example}.

\subsubsection{Weighing belief with heading measurement}
\label{ssec:heading_measurement}
We approximate the circular Gaussian presented in \secref{sec:heading_measurement} by von Mises distribution and we compute a weight matrix $\weightmatrix_{H,k}$ for all grid indices:
\begin{equation}
    \weightmatrix_{H,k}(i,j,l) =
    \Phi_{\mathcal{V}}(s_{\theta u}(l), v_k,1/\sigma_{v}^2) -
    \Phi_{\mathcal{V}}(s_{\theta v}(l), v_k,1/\sigma_{v}^2)
\end{equation}
where $\Phi_{\mathcal{V}}(t,\theta,1/\sigma^2)$ is the cumulative density function of von Mises distribution with parameters $\theta$, $1/\sigma^2$ evaluated at $t$.

\subsubsection{Weighing belief with map matching measurement}
For all grid indices, we compute a weight matrix $\weightmatrix_{M,k}$, from likelihood of the observation $\mathcal{I}_k$ representing the voxel $s_x(i), s_y(j), s_\theta(l)$, given map $\mathcal{M}$:
\begin{equation}
    \weightmatrix_{M,k}(i,j,l) = \cond{\mathcal{I}_k}{s_x(i), s_y(j), s_\theta(l),\mathcal{M}}
\end{equation}
The likelihood computation methods are described in detail in \secref{sec:determining_likelihood}.

\subsubsection{Updating with all measurements}
Our updated state estimate is computed as
\begin{equation}
    \boldx_{k,u} = \bar{\boldx}_{k,xy\theta} \odot \weightmatrix_{H,k} \odot \weightmatrix_{M,k},
\end{equation}
where $\odot$ is elementwise multiplication, and finally normalized:
\begin{equation}
    \boldx_{k} = \frac{\boldx_{k,u}}{\sum_{i,j,l}\boldx_{k,u}(i,j,l)}
\end{equation}
The end result is a recursive discrete approximation of equation \eqref{eq:state_estimate_continuous}.

\subsubsection{Interval for running algorithm}
Our localization algorithm is run at fixed intervals of travel, when the \ac{uav} has traveled more than a specified distance $u_{l}$ since the latest update. The amount of travel since latest update is approximated by odometry. The value for $u_{l}$ can be chosen to balance computational load, ensure enough movement with respect to to selected grid size, and have independence between \ac{uav} images used in map matching.

\section{Map matching}
\label{sec:map_matching}

Our formulation this far has considered how to fuse odometry, heading and map matching measurements using a point mass filter. We have still to define how to compute the likelihood $\cond{\mathcal{I}_k}{\continuousX_k,\mathcal{M}}$, \ie{} assess how well the observed \ac{uav} image $\mathcal{I}_k$ supports each possible value of state $\continuousX_k$, when using map $\mathcal{M}$.

Instead of detecting image features and using them as landmarks as done in \eg{} \cite{9311612}, we prefer an area-based approach, \ie{} using a large part of the observed image for discriminating between plausible and incorrect poses. The main motivation for this choice is that feature-based approaches require detection of spatially local features, which may be very sparsely detectable when flying over ambiguous terrains, especially across significant seasonal change (\eg{} after accumulation of snowfall) and when image footprint on ground is small. Using a large section of \ac{uav} image allows us to assess plausibility of matching with respect to a map even if locally distinct features cannot be detected.

Besides the targeted robustness when flying in ambiguous terrains, there are a number of other desirable characteristics for a map matching method. The method should work in matching patches of ground that are of a reasonable size; we should not develop a solution which requires a very large observed ground footprint and thus a high flying altitude in order for the solution to work. As we don't want to restrict the \ac{uav} trajectories, it is desirable to have a map matching method that is tolerant to viewpoint change (especially camera pitch and roll) and works at different flight altitudes and across a range of camera intrinsics. In addition, in developing a map matching approach trained with data samples, we cannot assume that a large amount of \ac{uav} imagery containing all types of expected variability (seasonal apperance change, camera angle with respect to ground and camera intrinsics, flight altitudes) is available for this purpose. Finally, the approach should be such that it works in the targeted scale and is computationally feasible for an onboard deployment. As a synthesis of all these needs, we propose an approach that splits the problem in two.

We first perform a rudimentary orthoprojection of the observed \ac{uav} image such that a section of the image that corresponds to a patch on ground of a specific size, at specific ground sampling distance, is generated (see \figref{fig:example_uav_image_and_projection_of_uav_image}). This acts as a means for abstracting away the flight altitude, camera intrinsic parameters and camera orientation with respect to ground and renders the problem of likelihood computation into that of finding likelihood between patches of orthoimages. This is beneficial also in training deep learning-based matching methods that are robust against seasonal variance as we demonstrated in earlier work \cite{9830867}, since our method abstracts away parameters that are irrelevant for the map matching problem (\ie{} camera intrinsics and altitude differences) and, unlike \ac{uav} image datasets, satellite images containing seasonal variation are plentiful.

As a second step, we compute a compact descriptor vector from the \ac{uav} image patch. We compute likelihood of each pose hypothesis by comparing that descriptor vector to a set of descriptor vectors that have been precomputed from map $\mathcal{M}$. Likelihood estimation is done by computing distance in the embedding space spun by the descriptor vectors. The choice of operating on compact descriptor vectors instead of template matching individual pose hypotheses as \eg{} in previous work \cite{9830867} is a key enabler for fast likelihood estimation over a large map, but raises the question of how to engineer a descriptor vector computation method in a way that allows as small vector size as possible while enabling robust localization.

In this section, we first introduce the orthoprojection method, after which we describe the methods of computing descriptor vectors, precomputing a map and finally computing the matching likelihood.

\subsection{Orthoprojecting \ac{uav} image}
\label{sec:orthoprojecting}

The map matching measurement is generated based on the view from a single \ac{uav} image, which may be tilted from nadir and thus contains perspective difference with respect to top-down view. To reduce the impact of this perspective change, we orthoproject the \ac{uav} image and make the assumption that the ground beneath the \ac{uav} is planar. Our approach resembles earlier work \cite{9659333} with a few key differences and we report an overview of the full approach for completeness.

Based on direction of gravity estimated by \ac{ahrs}, we first define local frame \localcoordinateframe{k} for image sampled at time $t_k$ whose origin is at the origin of the \ac{uav} camera frame, $z$ axis points in opposite direction to gravity, and the component of camera image plane horizontal axis perpendicular to $z$ is aligned with $y$ axis.

Our localization approach is based on using an orthoprojection of the camera view of the \ac{uav}. We detect image feature points using Shi-Tomasi detector~\cite{323794} and track movement of the features across a batch of ten consecutive camera image frames around keyframe sampled at $t_k$ using a pyramidal Lucas-Kanade tracker~\cite{bouguet2001pyramidal}. Besides image sampled at $t_k$, the batch consists of four images prior to the image corresponding with time $t_k$ and five after. We estimate the 3D locations of tracked feature points in frame \localcoordinateframe{k} using the linear triangulation method in \cite{HARTLEY1997146}, using relative pose transformations between frames in batch that we compute from ground truth data. Exploration of \ac{vio} frontends is beyond the focus of this work and we believe the use of noiseless relative transformations is a sufficient approximation of a generic \ac{vio} algorithm over the sequence of ten frames.

\begin{figure}
  \centering
  \begin{minipage}[t]{.45\linewidth}
    \subcaptionbox{Original \ac{uav} image $\mathcal{I}_k$}
      {\includegraphics[width=\linewidth]{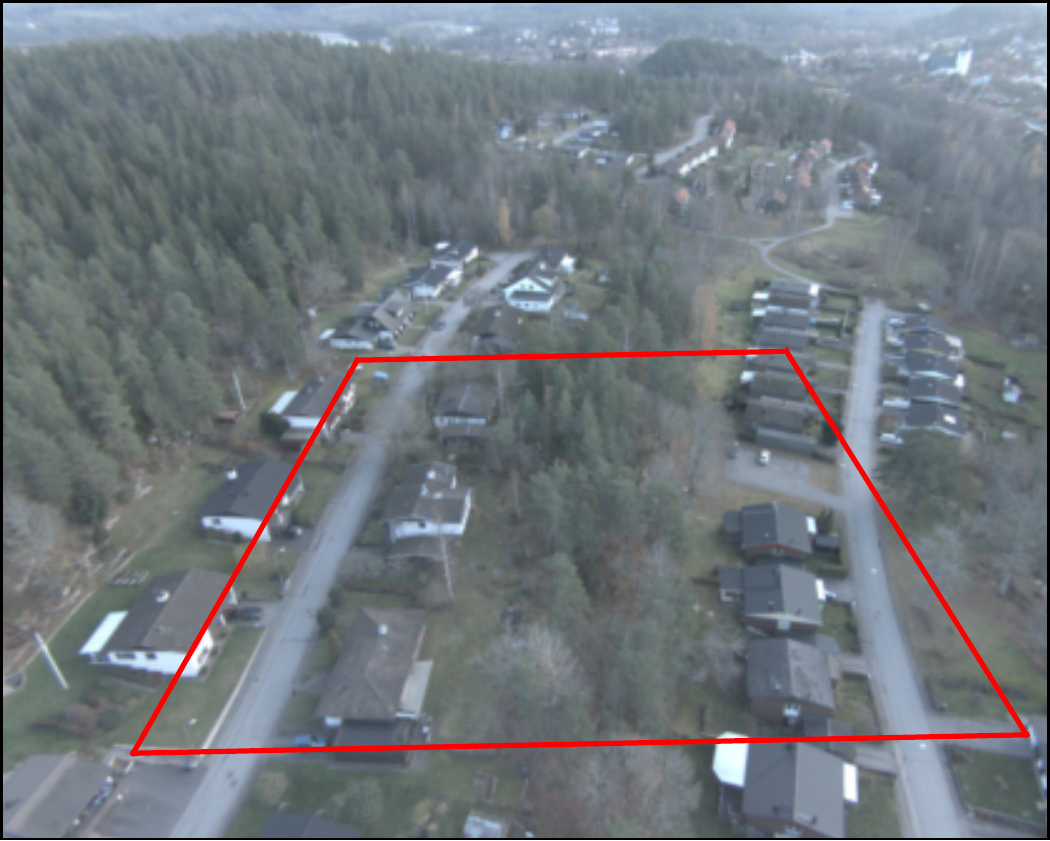}
      \label{fig:original_uav_image}
      }\\
    \subcaptionbox{Orthoprojected image $I_k$}
      {\includegraphics[width=\linewidth]{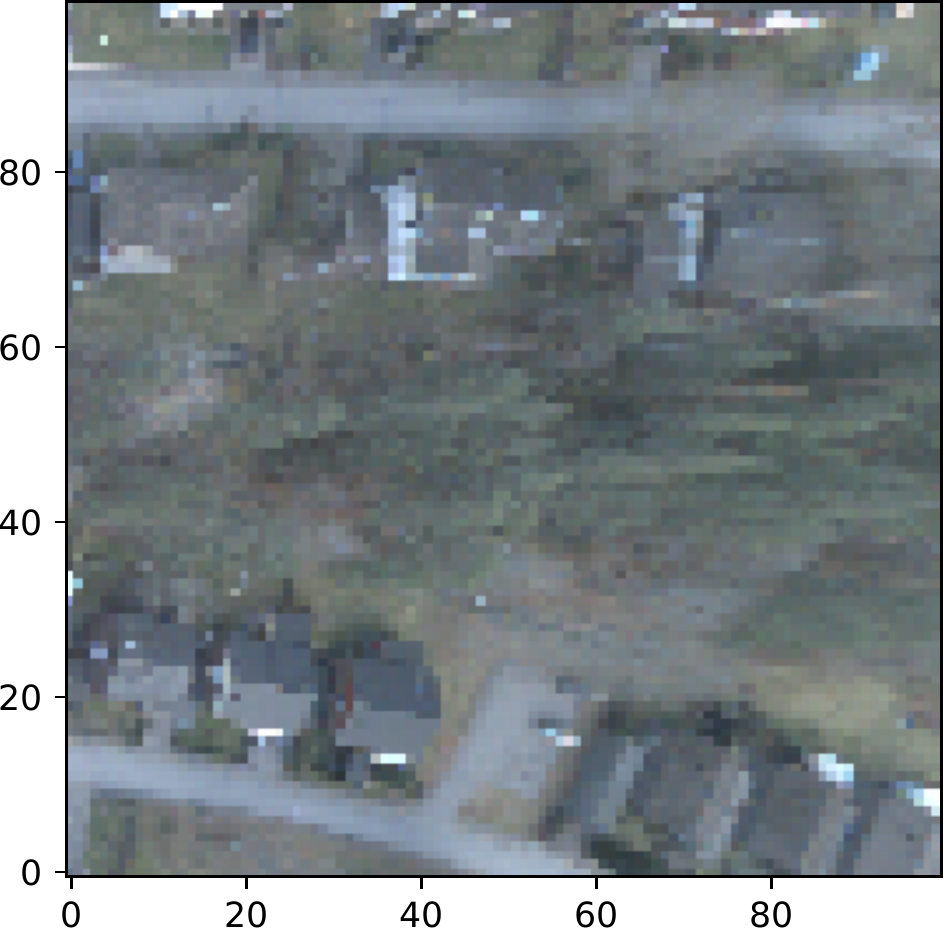}
      \label{fig:orthoprojected_uav_image}
      }%
  \end{minipage}%
  \hfill
  \begin{minipage}[b]{.50\linewidth}
    \subcaptionbox{Coordinate frame \localcoordinateframe{k}, points tracked in \ac{vio} (cyan dots), square lying on plane fit to \ac{vio} landmarks and transformation $\localcoordtosquarecoord{}$ to \squarecoordinateframe{k}}
      {\includegraphics[width=\linewidth]{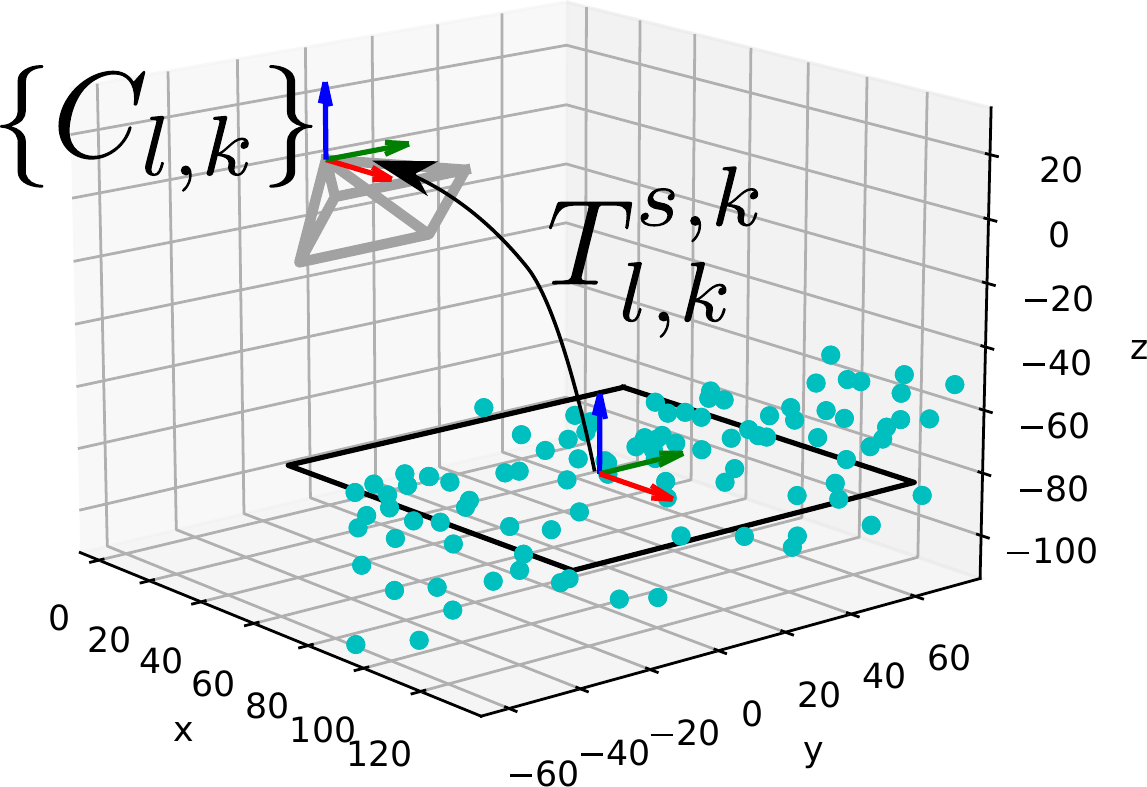}\label{fig:coordinate_frames}}%
  \end{minipage}%
  \caption
    {Example of \ac{uav} image (\squaresizem{} area used for orthoprojection highlighted in red), its orthoprojection, and a visualization of the coordinate frames.}
\label{fig:example_uav_image_and_projection_of_uav_image}
\end{figure}

We then find the best-fitting plane whose normal is aligned with $z$ axis; \ie{} we assume the ground below the \ac{uav} is planar and horizontal. We find a square with size \squaresizem{} lying on the best-fitting plane that is closest to nadir and fully visible in the camera image $\mathcal{I}_k$. We project the cornerpoints of this square into the image plane of $\mathcal{I}_k$ and, by homograpy, transform the pixels corresponding with the square in $\mathcal{I}_k$ to an \squaresizepx{} image at 1 m/px resolution which we call $I_k$. $I_k$ is the observation we use in map matching. We also compute $\localcoordtosquarecoord{}$, a transformation from frame \localcoordinateframe{k} to a frame centered in the middle of the square, which we label \squarecoordinateframe{k}. There is no rotation between \localcoordinateframe{k} and \squarecoordinateframe{k} and translation is defined such that \squarecoordinateframe{k} is at the center of the square. An example of one orthoprojected \ac{uav} image and a visualization of the coordinate frames is shown in \figref{fig:example_uav_image_and_projection_of_uav_image}.

The presented approach enables us to consider the motion of a \ac{uav} as a sequence of 2D translations and rotations while allowing extraction of observation $I_k$ at correct scale independent of fight altitude, while $\localcoordtosquarecoord{}$ contains information about position of its centerpoint relative to camera, including estimated ground plane altitude.

\subsection{From orthoimage patch to descriptor vector}
\label{sec:observation_into_embeddingspace}
Inspired by other works projecting the \ac{uav} observation into a single compact descriptor vector \cite{9357892, 9562005, 10.1117/12.2585986}, to provide a fast way to compare an observation to a large number of pose hypotheses, our map matching method is based on transforming both \ac{uav} observations and a reference map into a suitable descriptor space, and computing probability of pose hypotheses using descriptor vector values.
Based on image $I_k$, we compute a descriptor vector $\descriptorvector_k$ using function $f_i$: $\descriptorvector_k = f_i(I_k)$. This function $f_i$ is composed of a deep neural network backbone $b$ such as Resnet \cite{He_2016_CVPR} followed by a projection module $m$. Therefore $f_i(I_k)= m_i(b_i(I_k))$. The projection module is stacked on top the ResNet model where the the last average pooling and final fully connected layer specified in \cite{He_2016_CVPR} are removed. The last layer of the projection module has $D$ neurons in order to compute $D$-dimensional $l_2$-normalized descriptor vectors. The projection model thus produces vectors in unit $D$-sphere. Therefore, $f: \mathbb{R}^{h \times w \times c} \rightarrow \mathbb{R}^{D}$, where $h$, $w$ and $c$ are the height, width and channels dimensions of the input image $I_k$.
\subsubsection{Projection modules}
\label{sssec:projection_modules}

In order to extract $D$-dimensional embedding vectors, we used two different types of projection modules: one composed only of fully connected layers, $m_{FCN}$, and one composed by a Capsule Network (CapsNets) model~\cite{hinton-dynamic}, $m_{CAP}$. We refer to $f_{FCN}$ and $f_{CAP}$ as the models where the ResNet backbone is followed by $m_{FCN}$ and $m_{CAP}$, respectively.

\begin{figure}
     \centering
     \begin{subfigure}[b]{0.7\linewidth}
         \centering
         \includegraphics[width=\textwidth]{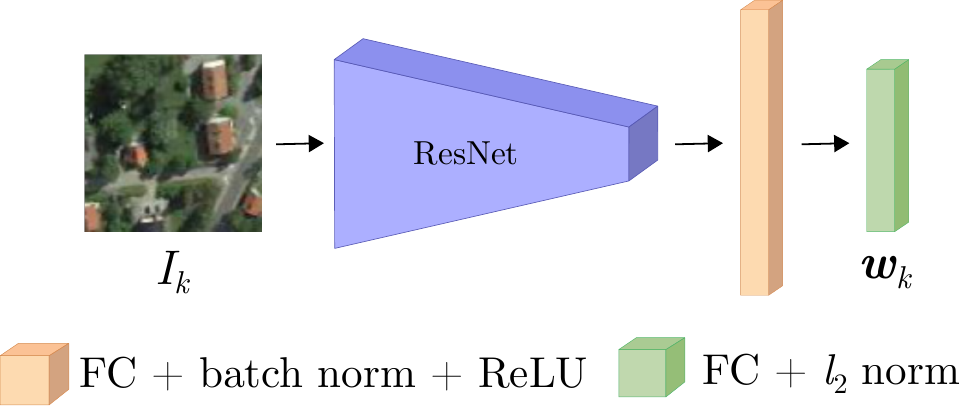}
         \caption{$f_{FCN}$}
         \label{fig:fcn_architecture}
     \end{subfigure}
     \par\bigskip
     \begin{subfigure}[b]{\linewidth}
         \centering
         \includegraphics[width=\textwidth]{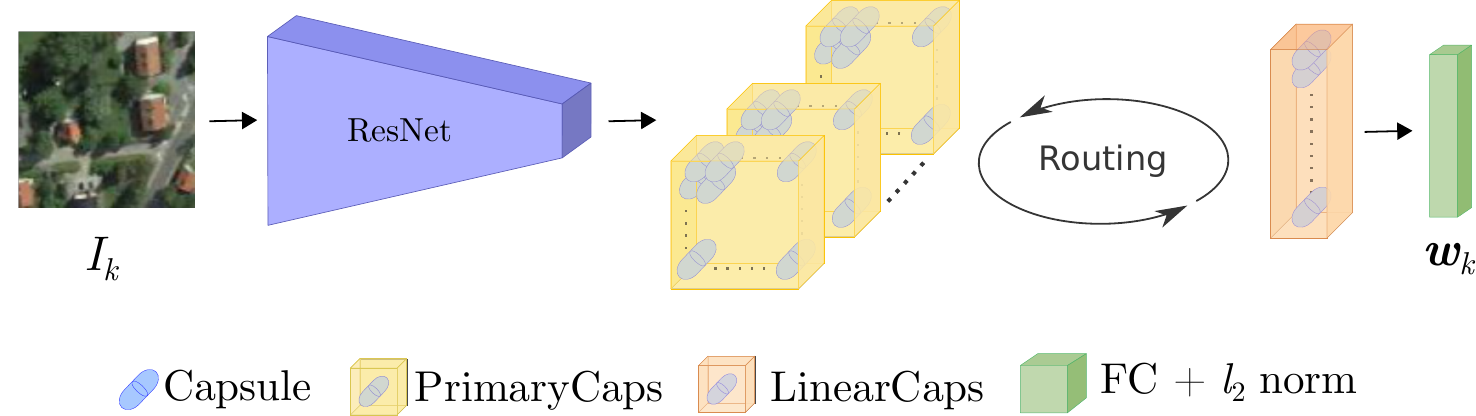}
         \caption{$f_{CAP}$}
         \label{fig:caps_architecture}
     \end{subfigure}
     \hfill     
        \caption{Architectures for fully connected and capsule networks}
        \label{fig:f_architectures}
\end{figure}

Inspired by the architecture choices proposed by \cite{9562005,10.1117/12.2585986}, the $m_{FCN}$ module is composed by two fully connected layers, one with N neurons and one with D outputs. A visualization of the network structures is shown in \figref{fig:fcn_architecture}.

CapsNets have gained great attention recently since they are more robust to input perturbations compared to other \ac{cnn} architectures with a similar number of trainable parameters~\cite{hinton-dynamic, hinton-em}. Their main innovation lies in two major distinctions from the \acp{cnn}:
(i) the encoding of object poses (position, size, orientation) and visual attributes (\eg{} color, texture, deformation, hue) into groups of neurons called \textit{capsules} (ii) the routing-by-agreement mechanism, which models the connections between capsules of different layers. Namely, it models the part-whole relationships among objects without losing spatial information.
We explore the use of CapsNets as projection modules for two main reasons: first, CapsNets are well known for being robust to affine transformations and novel viewpoints; secondly, they have also been shown to achieve higher generalization with less data compared to CNNs~\cite{hinton-dynamic, hinton-em}. Owing to these properties, extracting descriptor vectors using $f_{CAP}$ can help to solve the \ac{uav} localization task described in this work. Furthermore, in the literature~\cite{self-routing, gracapsnets}, capsule layers have been widely stacked on top of ResNet backbones instead of traditional fully connected layers to achieve better performance. A visualization of $f_{CAP}$ is shown in \figref{fig:caps_architecture}.

\subsection{Precomputing descriptors for map}
\label{ssec:precomputing_map}
We precompute offline a map around the expected operating region. The map holds descriptor vector values that have been computed from a georeferenced orthophoto RGB bitmap $\mathcal{M}$. For each map coordinate $X_h = (x_h, y_h, \theta_h)$, we crop a $w$ m by $w$ m square image $I_{\mathcal{M},h}$, translated from origin by $(x_h, y_h)$ in map coordinates and rotated by $\theta_h$, from the map image $\mathcal{M}$ and scale it to 1 m/px resolution. We then compute an embedding vector $\descriptorvector_h = f_i(I_{\mathcal{M},h})$. We compute these embedding vectors using the centerpoint coordinate of each voxel in $\boldx$. This process yields a precomputed map $\precomputedmap(i,j,l) \in \mathbb{R}^{D}$ where indices $i$, $j$, $l$ correspond with indices of grid cells in our belief representation $\boldx$. We compute a separate map for each tested network architecture.

\subsection{Determining map matching likelihood}
\label{sec:determining_likelihood}
To assess whether a \ac{uav} observation corresponds with a state hypothesis, we approximate
\begin{equation}
    \cond{\mathcal{I}_k}{s_x(i), s_y(j), s_\theta(l),\mathcal{M}} \approx \cond{\descriptorvector_k}{i,j,l,\precomputedmap}
\end{equation}
and compute $\weightmatrix_{M,k}$, which contains the weight for each state space element on the chosen voxel grid.
\begin{equation}
    \weightmatrix_{M,k}(i,j,l) = \cond{\descriptorvector_k}{i,j,l,\precomputedmap}
\end{equation}

We compare two solutions for this. The first one, labeled \emph{linear}, is similar to the choice in \cite{9562005}. It assumes that Euclidean distance between the embedding obtained from the \ac{uav} and from the map $c_k(i,j,l) = \|\descriptorvector_k-\precomputedmap(i,j,l)\|_2$ is inversely proportional to the probability of correct pose:
\begin{equation}
    \weightmatrix_{M1,k}(i,j,l) =
    \frac{2-c_k(i,j,l)}{2}
\end{equation}

The second one is the method for computing importance factor presented in earlier work \cite{9830867}, where we estimate the probability density of distances in Euclidean space for true and false matches from satellite image data and compute the probability that the observation is from "match" class, for each element in $\boldx_k$ individually. We label this weighing method \emph{bayesian} and we name weight matrix $\weightmatrix_{M2,k}$.

\section{Experiments}\label{sec:experiments}

\subsection{Overview of experiments}

We experiment the performance of our solution with respect to baseline methods on real-world datasets. We evaluate probabillity of convergence, time to convergence and localization error after convergence with flights taking place in two areas in Sweden. In both areas, we experiment with the problem of localization starting from \maparea{} uncertainty. In addition, we experiment with real-time implementation on a \ac{uav} onboard computer.

For evaluating the critical design choices in our map matching approach, we vary the projection module type (fully connected or capsule network), likelihood vector dimensionality $D$ (8, 16, 32 or 128) and likelihood conversion method (linear or bayesian), evaluate the impact of these choices on probability of convergence, time to convergence and mean localization error after convergence.

We evaluate localization performance by the criteria defined in \secref{sec:evaluation_criteria_in_experiments}. The datasets we use for experimentation and model training are described in \secref{ssc:datasets}. Training methods are outlined in \secref{subsubsec:training_details}, baseline methods are described in \secref{ssec:comparisonmethods}, followed by evaluation of localization performance in \secref{ssec:evaluating_localization_performance} and learnings from real-time experiments in \secref{sec:realtime_experiment}.

\subsection{Evaluation criteria in localization experiments}
\label{sec:evaluation_criteria_in_experiments}
\subsubsection{Translation error}

We compute the estimated \xycoords~ and heading using \eqref{eq:xy_estimate} and \eqref{eq:theta_estimate}, respectively.
\begin{subequations}
\begin{equation}
    \label{eq:xy_estimate}
    \widehat{X}^{xy}_{s,k} =
    \sum\limits_{i,j,l}\boldx(i,j,l)
    \begin{bmatrix}
    \bar{s}_{x}(i) & \bar{s}_{y}(j)
    \end{bmatrix}^T
\end{equation}
\begin{equation}
    \label{eq:theta_estimate}
    \widehat{X}^{\theta}_k = atan2(\sum\limits_{i,j,l}\boldx(i,j,l)
    sin(\bar{s}_{\theta}(l)),\sum\limits_{i,j,l}\boldx(i,j,l)
    cos(\bar{s}_{\theta}(l)))
\end{equation}
\end{subequations}
Here, $\bar{s}_x(i)$, $\bar{s}_y(j)$ and $\bar{s}_\theta(l)$ are the centerpoint coordinates of voxel corresponding to indices $(i, j, l)$ and $atan2$ is the 2-argument arctangent. We then transform these estimates from $\widehat{X}^{xy}_{s,k}$ to drone-centric coordinates $\widehat{X}^{xy}_{k}$.

We compute the Euclidean distance to \ground{} \xycoords{} $X^{xy}_{k,gt}$:

\begin{equation}
    \label{eq:xy_error}
    \widetilde{X}^{xy}_k = \| \widehat{X}^{xy}_k - X^{xy}_{k,gt} \|_2
\end{equation}

We also compute
$\sigma_{\widehat{xy}}$, the weighted standard deviation of Euclidean distances to $\widehat{X}^{xy}_k$ in $(x,y)$ plane, weighing with $\boldx_k$. We claim that $\sigma_{\widehat{xy}}$ is a suitable measure of convergence of the localization solution, and a large spread of uncertainty signals the need for re-initialization of the estimator.

\subsection{Datasets}\label{ssc:datasets}
\subsubsection{Satellite images}\label{subsubsec:sat_dataset}
We trained our networks on 100m by 100m samples at 1 m/pixel randomly drawn from Google Earth satellite images of size 4800 by 2987 meters collected from 9 regions, in arbitrarily selected places in southern Finland that cover urban and non-urban areas. For each region, 4 to 15 satellite images collected from the same area at different times, containing seasonal variation, are used. The Google Earth datasets are the same as in an earlier work \cite{9830867}.
\subsubsection{UAV images}\label{subsubsec:uav_dataset}
\begin{figure}
     \centering
     \begin{subfigure}[b]{0.22\linewidth}
         \centering
         \includegraphics[width=\textwidth]{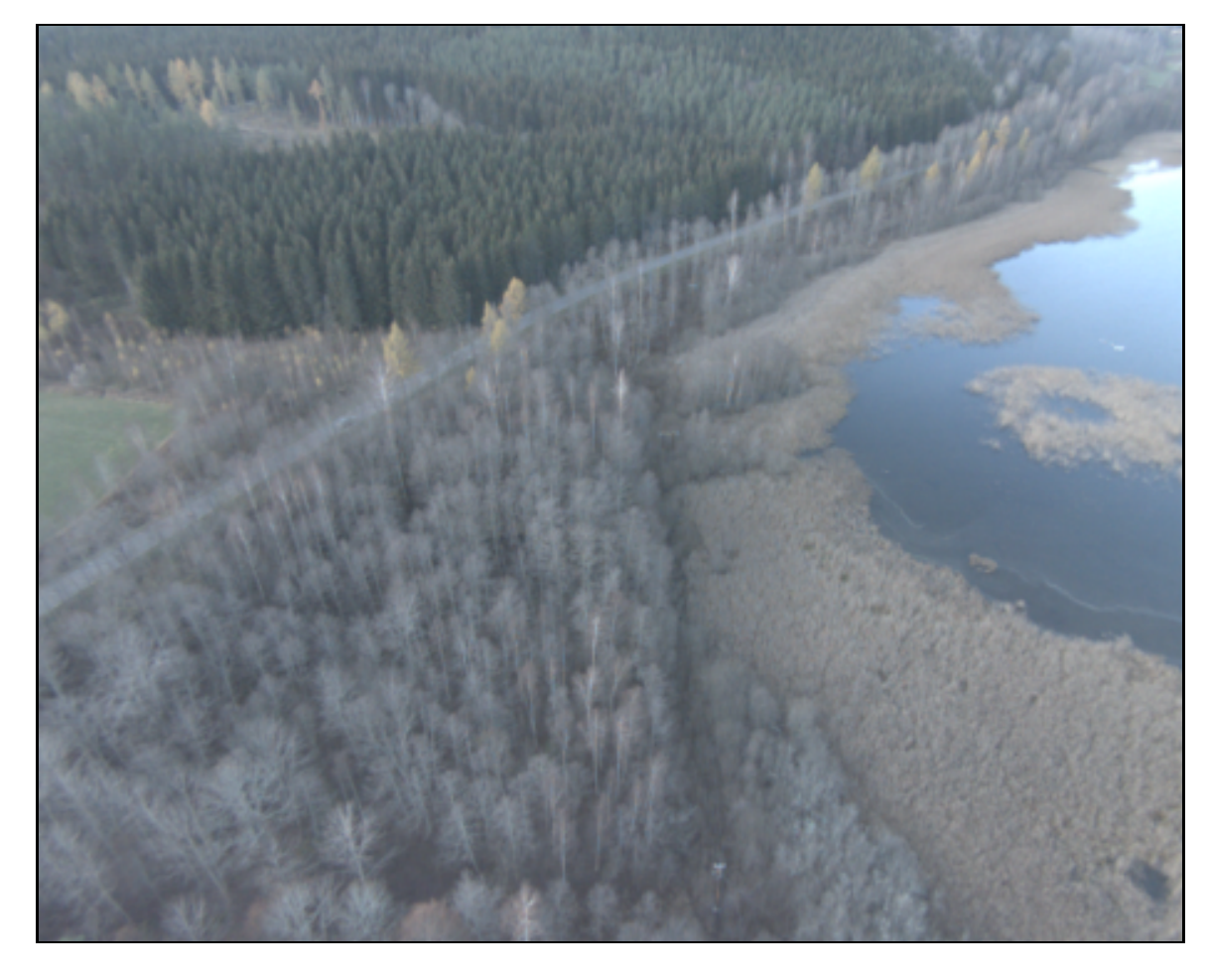} 
         \caption{Dataset 1}
         \label{fig:sampleimages_dataset1}
     \end{subfigure}
     \hfill
     \begin{subfigure}[b]{0.22\linewidth}
         \centering
         \includegraphics[width=\textwidth]{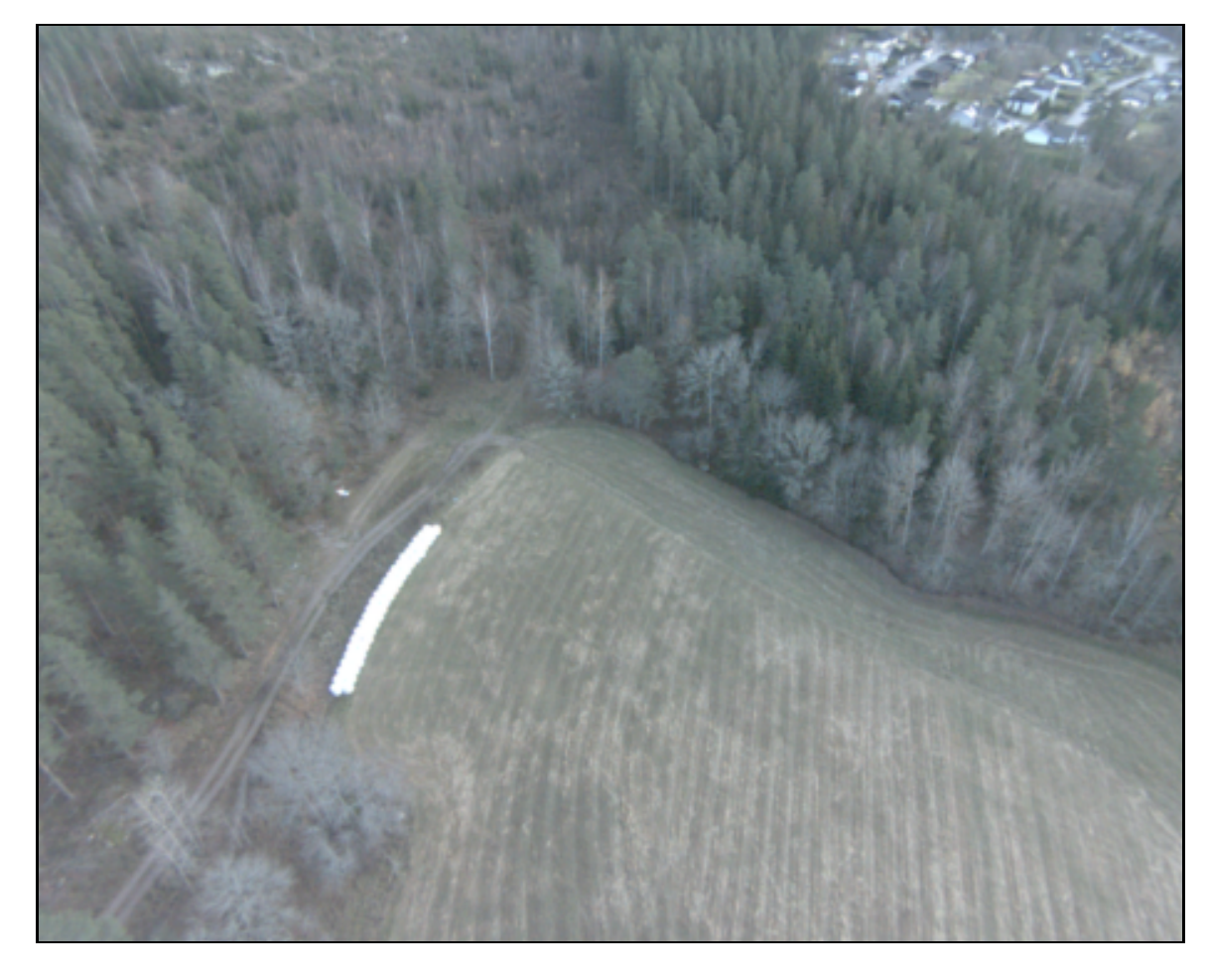} 
         \caption{Dataset 2}
         \label{fig:sampleimages_dataset2}
     \end{subfigure}
     \hfill
     \begin{subfigure}[b]{0.22\linewidth}
         \centering
         \includegraphics[width=\textwidth]{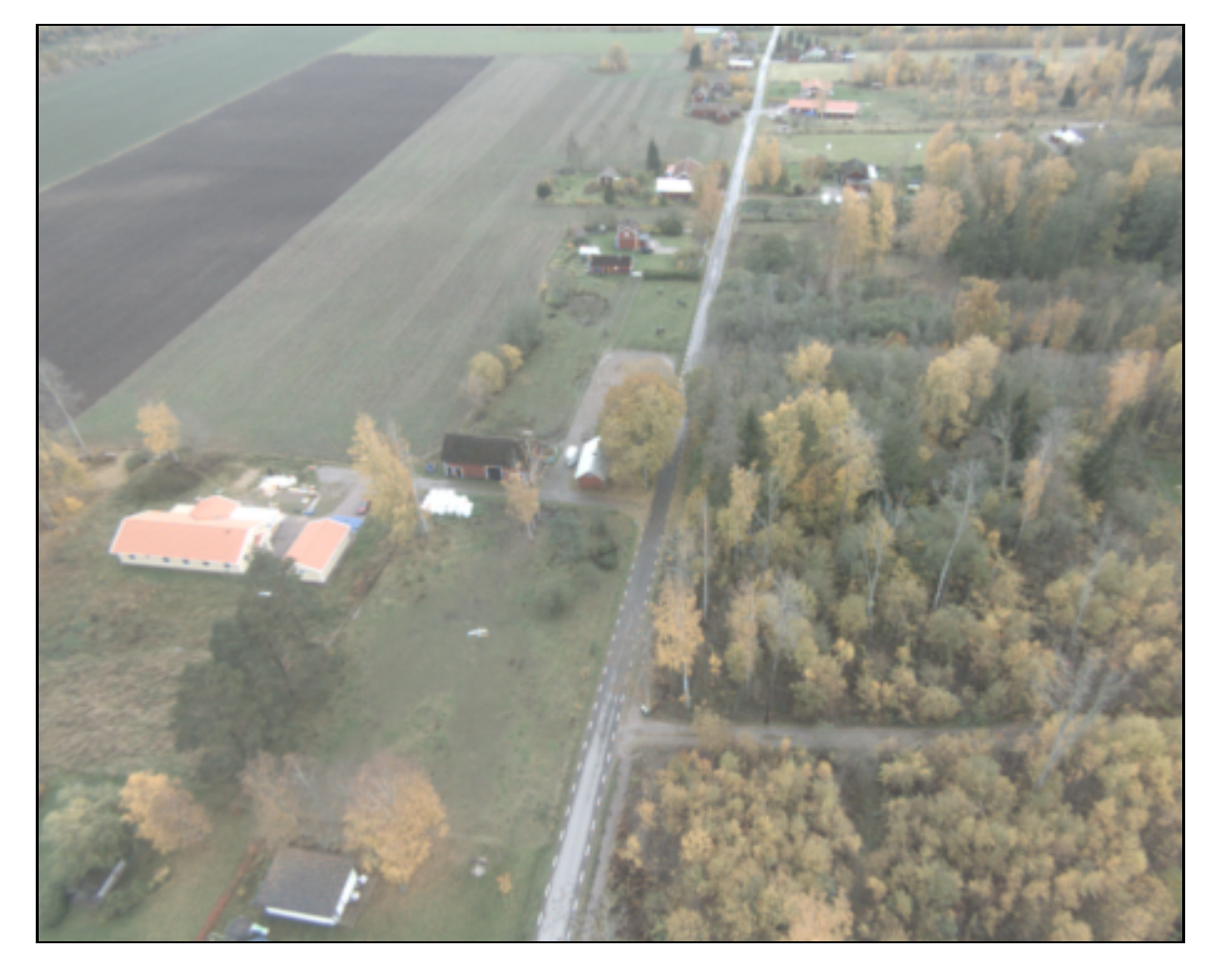} 
         \caption{Dataset 3}
         \label{fig:sampleimages_dataset3}
     \end{subfigure}
     \hfill
     \begin{subfigure}[b]{0.22\linewidth}
         \centering
         \includegraphics[width=\textwidth]{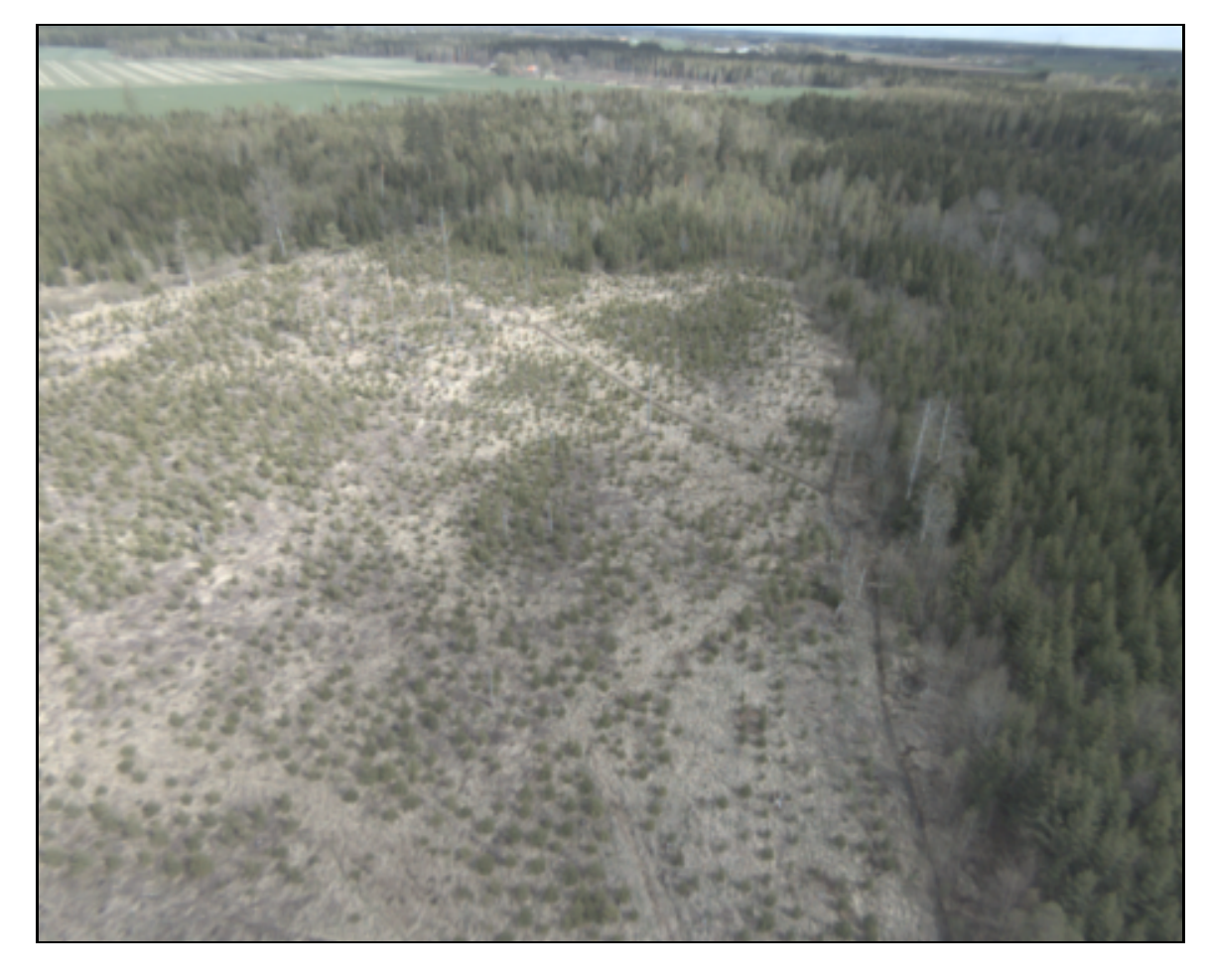} 
         \caption{Dataset 4}
         \label{fig:sampleimages_dataset5}
     \end{subfigure}     
     \hfill
     \begin{subfigure}[b]{0.22\linewidth}
         \centering
         \includegraphics[width=\textwidth]{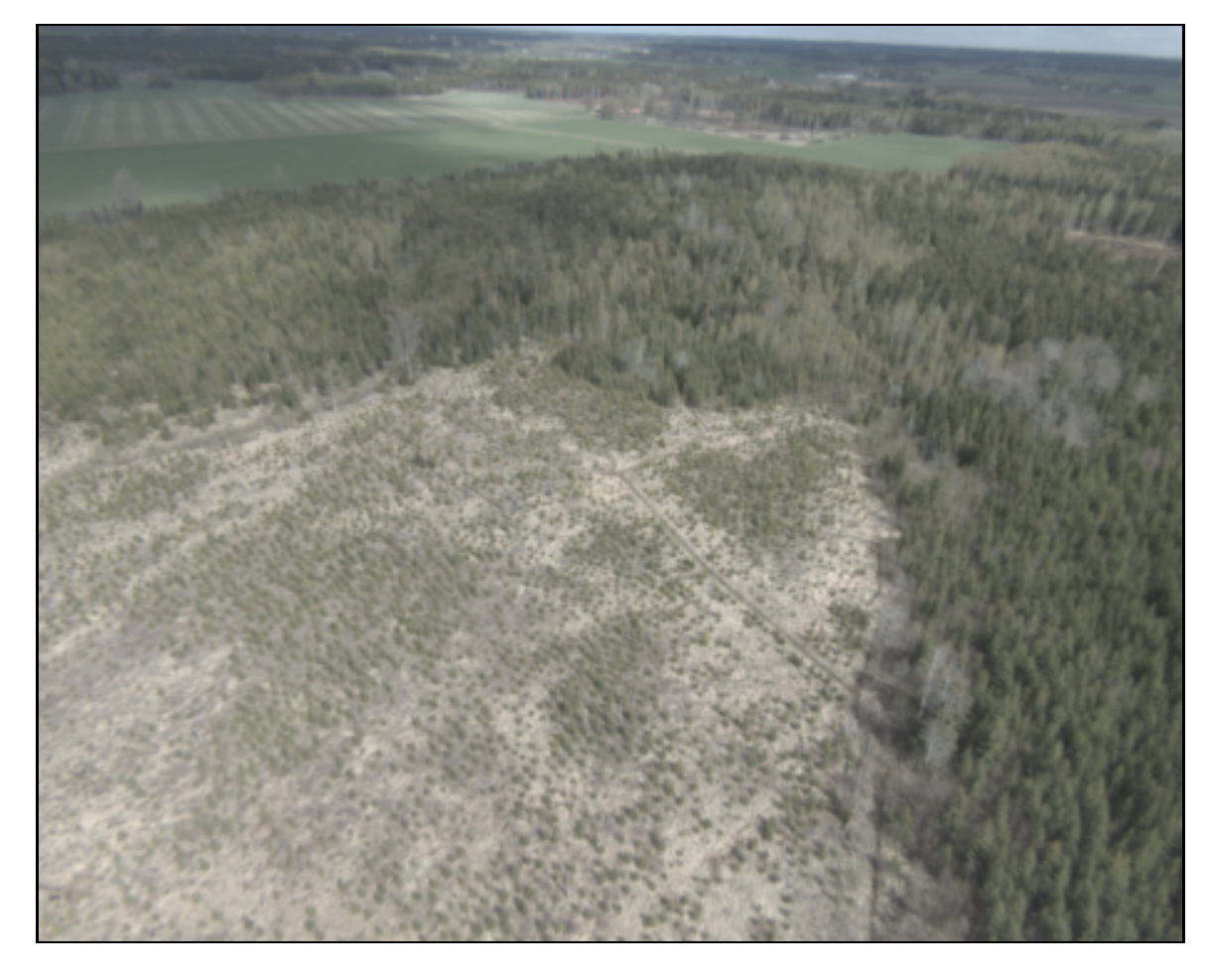} 
         \caption{Dataset 5}
         \label{fig:sampleimages_dataset6}
     \end{subfigure}
     \hfill
     \begin{subfigure}[b]{0.22\linewidth}
         \centering
         \includegraphics[width=\textwidth]{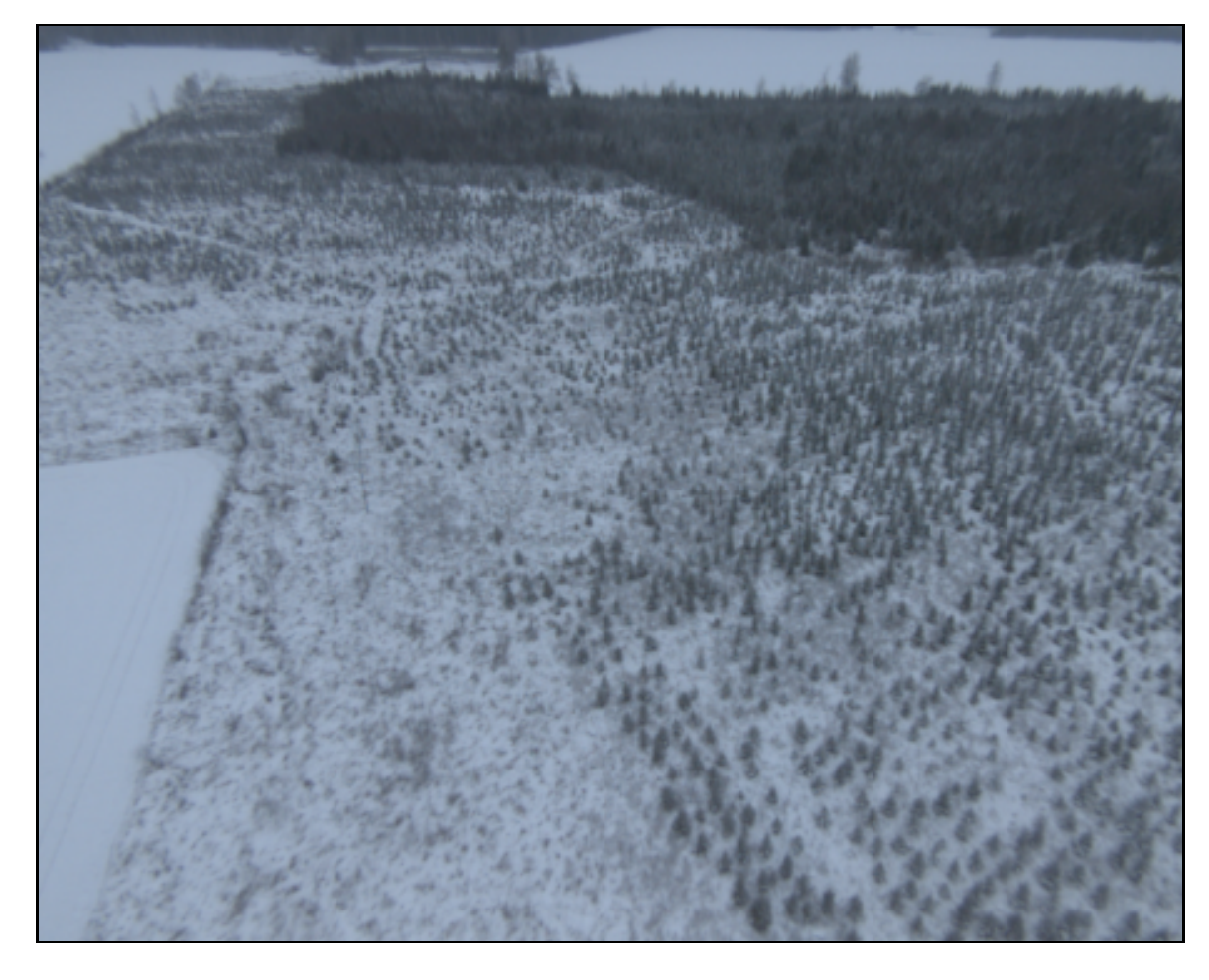} 
         \caption{Dataset 6}
         \label{fig:sampleimages_dataset7}
     \end{subfigure}
     \hfill
          \begin{subfigure}[b]{0.22\linewidth}
         \centering
         \includegraphics[width=\textwidth]{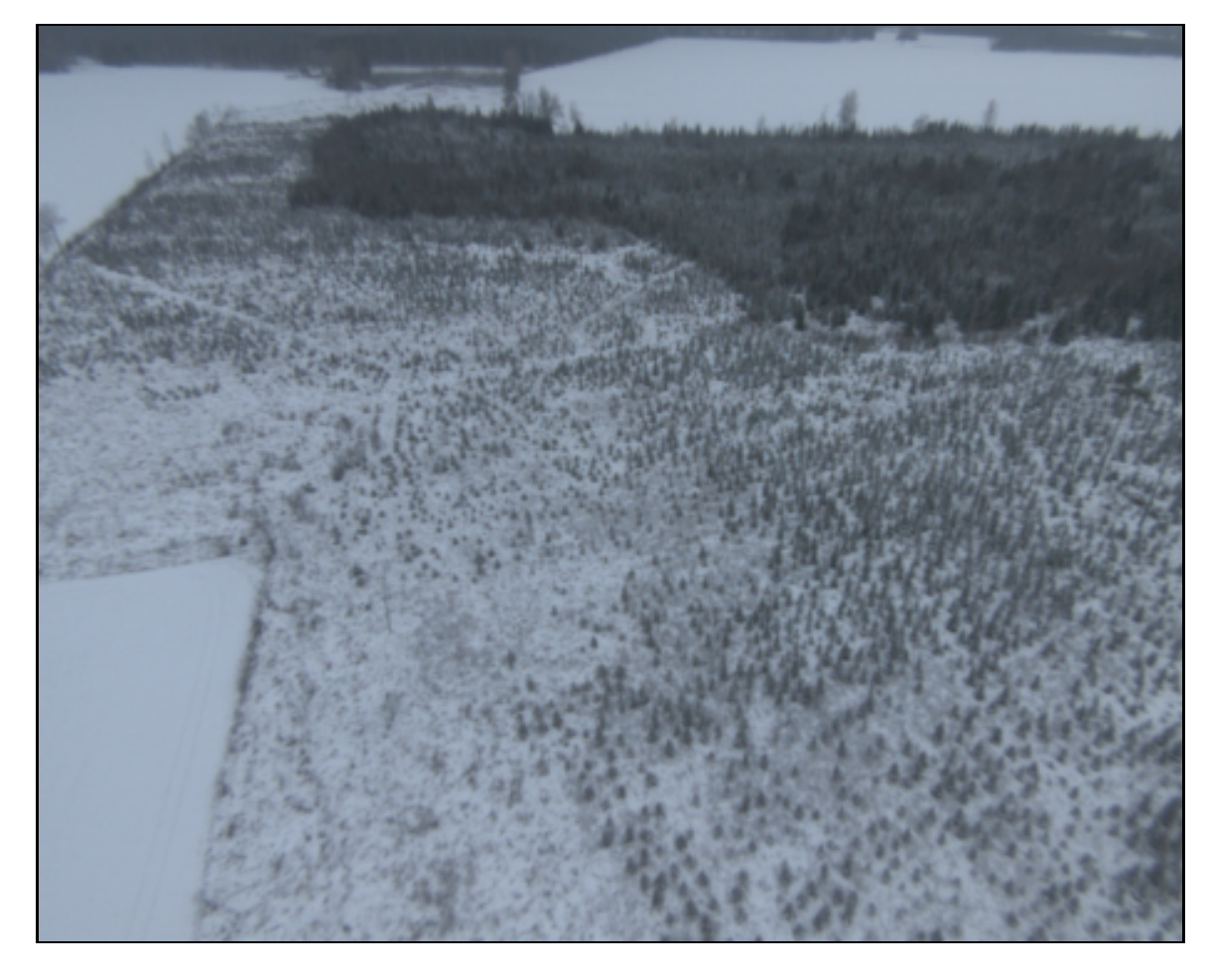} 
         \caption{Dataset 7}
         \label{fig:sampleimages_dataset4}
     \end{subfigure}
     \hfill
     \begin{subfigure}[b]{0.22\linewidth}
         \centering
         \includegraphics[width=\textwidth]{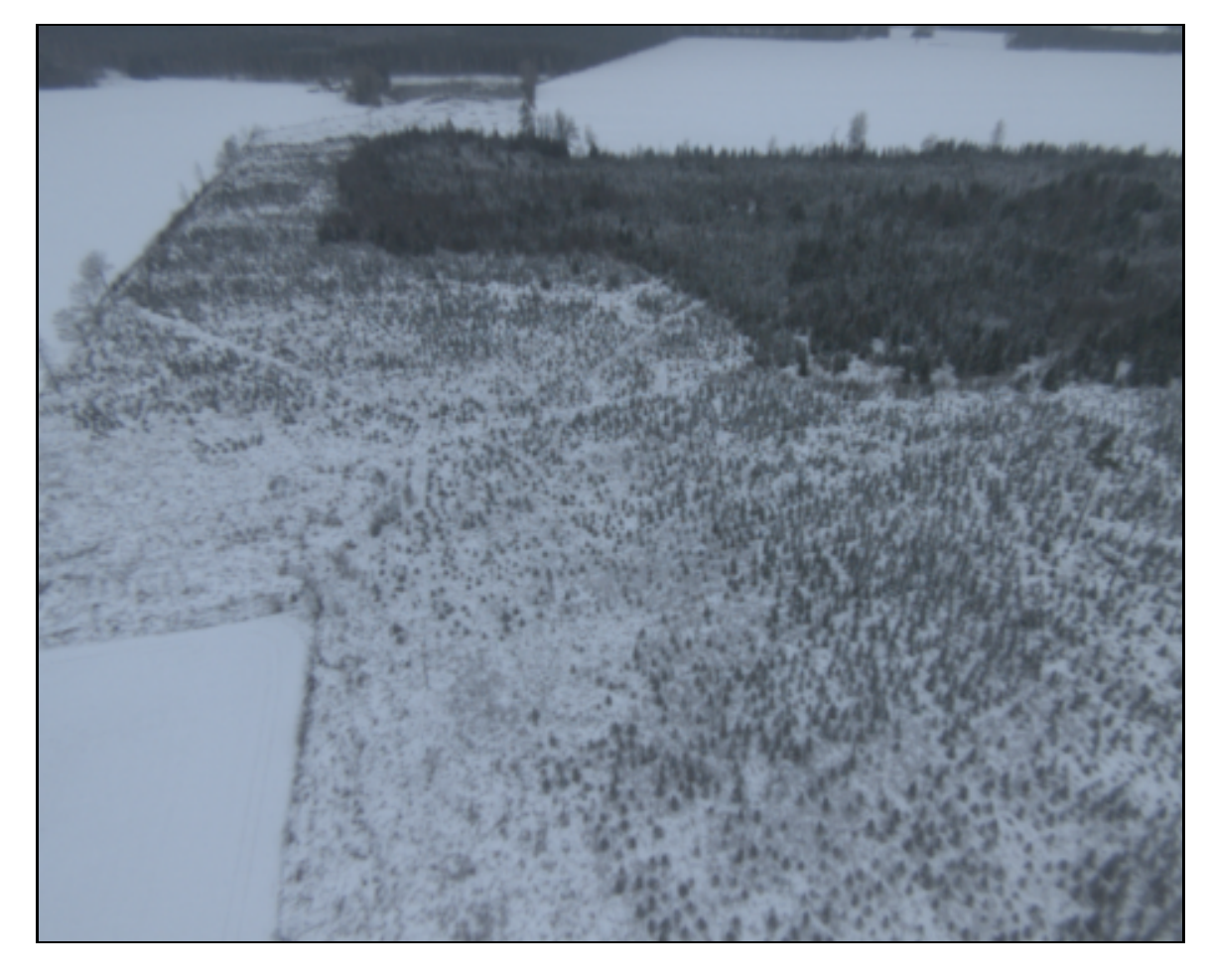} 
         \caption{Dataset 8}
         \label{fig:sampleimages_dataset8}
     \end{subfigure}
     \hfill     
        \caption{Example images from datasets. Note difference in seasonal appearance.}
        \label{fig:example_images_from_datasets}
\end{figure}

\begin{figure}
     \centering
     \begin{subfigure}[b]{0.49\linewidth}
         \centering
         \includegraphics[width=\textwidth]{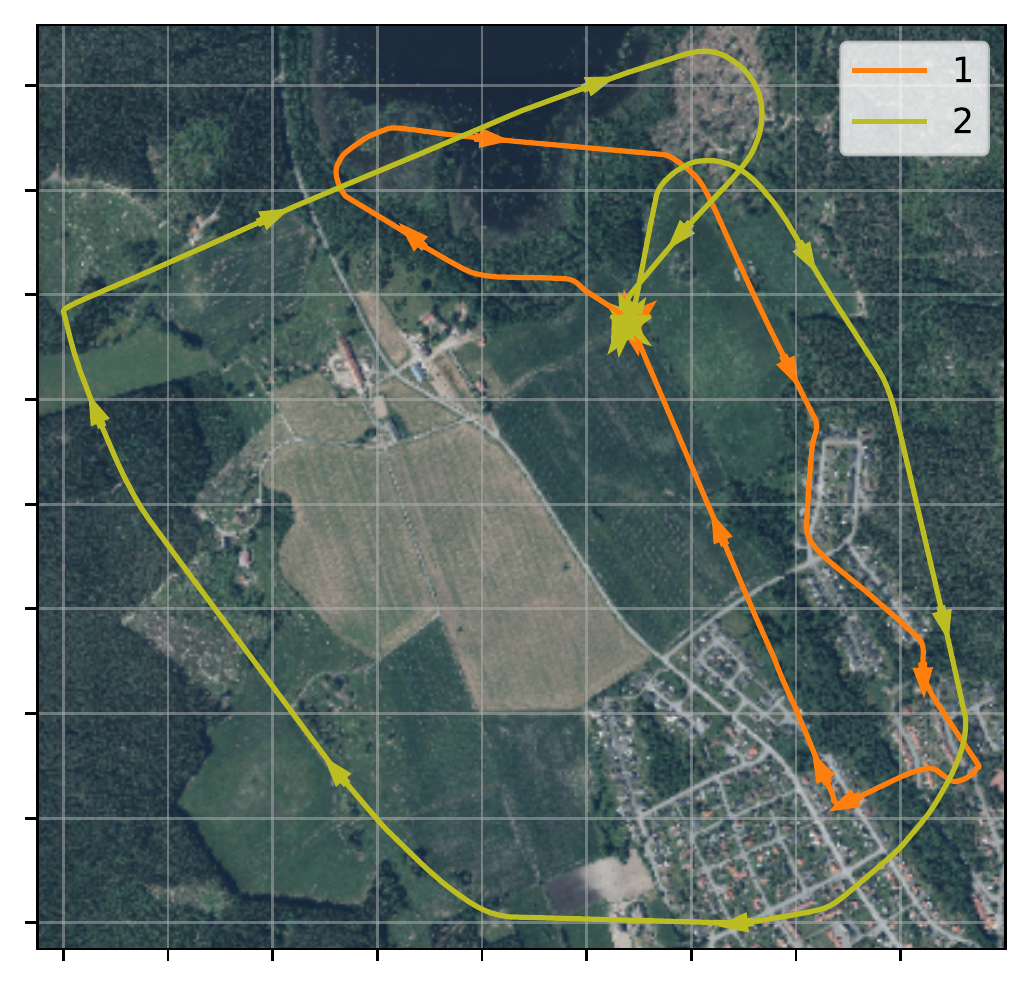}
         \caption{Kisa}
         \label{fig:area_a_trajectories}
     \end{subfigure}
     \hfill
     \begin{subfigure}[b]{0.3\linewidth}
         \centering
         \includegraphics[width=\textwidth]{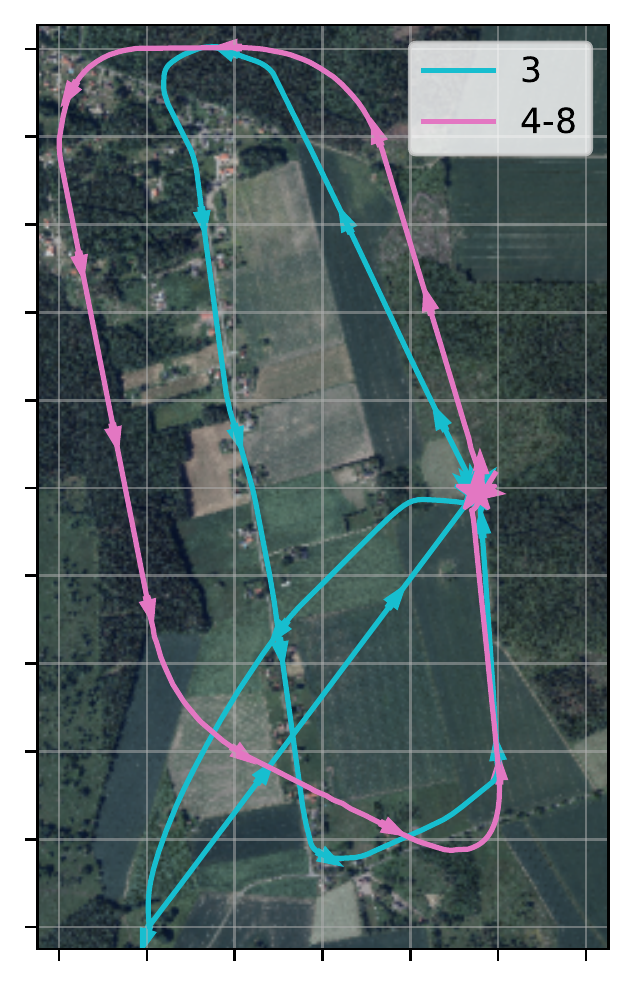}
         \caption{Klockrike}
         \label{fig:area_b_trajectories}
     \end{subfigure}
     \hfill     
        \caption{Flight trajectories of each dataset. Grid spacing 200 m. Datasets 4-8 have same trajectory with varying altitude and season.}
        \label{fig:trajectories}
\end{figure}

Our experiments are run on datasets that have been collected with a \ac{uav} in two locations in Sweden at different times\footnote{Data provided by Saab Dynamics Ab.}. The dataset contains posed images sampled at 10 Hz. \Ground{} camera poses are fused from \ac{rtk}-corrected \ac{gps} in uninterfered conditions and \ac{imu} measurements with a proprietary algorithm. The use of RTK-GNSS ensures position precision in the centimeter range. A listing of the flight experiments is given in \autoref{tab:uav_dataset_characteristics}. A representative image of each dataset is shown in \autoref{fig:example_images_from_datasets} and flight trajectories over an orthophoto map are shown in \autoref{fig:trajectories}. The flights take place over terrains with forest areas, a lake, agricultural fields and some residential areas. At the time of running this experiment, the original \ac{imu} data was not available.

\subsubsection{Maps used for localization}
Each original map $\mathcal{M}$ is an orthophoto bitmap constructed from aerial images taken over the operating area in summer 2021. We use orthophoto bitmaps with an original ground sampling distance of 0.16 m/px provided by a local map information supplier\footnote{\copyright ~Lantmäteriet, \url{https://www.lantmateriet.se/}.}.

\begin{table}
\caption{\label{tab:uav_dataset_characteristics}Characteristics of flight datasets. Trajectory lengths are computed along $(x, y)$ plane, and camera angles between nadir and camera principal axis. Altitude is with respect to starting position.}
\begin{tabular}{p{2mm} p{10mm} p{14mm} p{6mm} p{4mm} p{22mm}} 
\toprule
\# & Area & Date & Traj. length (km) & Alt. (m) & Median camera angle [range] ($^{\circ}$)\\
\midrule
1 & Kisa & 2019-11-07 & 4.1 & 91 & 59.0 [56.9 \ldots 67.7] \\ 
2 & Kisa & 2019-11-07 & 6.1 & 92 & 52.1 [49.8 \ldots 56.0] \\ 
3 & Klockrike & 2019-10-18 & 6.8 & 92 & 51.2 [49.9 \ldots 53.9] \\ 
4 & Klockrike & 2020-04-29 & 4.8 & 53 & 59.3 [57.8 \ldots 60.5] \\ 
5 & Klockrike & 2020-04-29 & 4.8 & 92 & 59.9 [58.4 \ldots 61.3] \\ 
6 & Klockrike & 2021-01-19 & 4.9 & 54 & 58.9 [58.0 \ldots 60.4] \\ 
7 & Klockrike & 2021-01-19 & 4.7 & 71 & 58.7 [57.7 \ldots 60.0] \\ 
8 & Klockrike & 2021-01-19 & 4.9 & 84 & 58.4 [56.9 \ldots 60.9] \\ 
\bottomrule
\end{tabular}
\figvspace{}
\end{table}
\subsection{Training descriptor generator networks}

\subsubsection{Training details}\label{subsubsec:training_details}
We trained the models $f_{FCN}$ and $f_{CAP}$ for 500 epochs only on satellite images.\footnote{We acknowledge the computational resources provided by the Aalto Science-IT project.} At each epoch, 45k training satellite samples are generated from the 9 areas. Each sample represents a location and orientation on a map. We draw 5 images corresponding with each location and orientation. For each batch, we randomly select 10 locations. As deep metric learning for encoding the images into low dimensional vectors, we employed the triplet loss~\cite{tripletloss} with batch-all strategy and margin $0.2$, using vectors representing the same location and orientation as positive samples, and vectors representing other locations and orientations in the batch as negative samples. In addition to random selection of locations and orientations, we add a random translation offset of 0 to 35 meters with uniform distribution to each location, with the intent to add robustness for observations that do not align perfectly on the map grid. With similar motivation, we add normally distributed random rotations with standard deviation of 6 degrees to the orientation of each sample before extracting the image patch from the satellite image. In addition to these, we augment the samples with Gaussian noise, motion blur, random brightness and contrast changes, and hue and saturation changes to add tolerance for changes in imaging conditions. We used Adam optimizer~\cite{adam} with learning rate $10^{-6}$. 

\subsubsection{Architecture details of neural networks}
We employed a Resnet-50 model implemented in PyTorch 1.8.0 as backbone $b$. This network is pretrained on Places365 dataset~\cite{zhou2017places}. As regards the projection module $m_{FCN}$, we used a fully connected layer of sizes 1024 and $D$, respectively. We experiment with vector length $D \in \{8, 16, 32, 128\}$. On the other hand, with regards to the module $m_{CAP}$, the PrimaryCaps layer is a capsule layer with 64 types of 16-dimensional capsules, which are obtained from a convolutional layer of 1024 $1\times1$ kernel size filters. The LinearCaps layer consists of 32 32-dimensional capsules extracted running 3 iterations of the routing algorithm. We stacked on top of the LinearCaps layer a fully connected layer of size $D$. With the capsule projection module, we explore $D \in \{8, 16, 32\}$.
For both projection modules, the $D$-dimensional output of the last layer is used to produce the $l_2$-normalized embedding vectors.

\subsection{Comparison methods}\label{ssec:comparisonmethods}

We implement two methods to work as comparison methods of map matching for our approach. To provide comparable results, we utilize the same point mass filter implementation, same odometry measurements, prediction method, heading weighing method and same measurement images $I_k$ for all methods and only replace the map matching solution with their approach.

As baseline methods for map matching, we use the solution by Mantelli \etal{} \cite{MANTELLI2019304}, whose formulation is scalable to large maps. To provide comparable results, we compute the descriptor vectors $\descriptorvector_k = f_i(I_{k})$ such that $\descriptorvector_k$ is the abBRIEF descriptor. In a similar manner, we precompute a grid of abBRIEF descriptors from $\mathcal{M}_b$, using equal grid spacing as with our methods, and with abBRIEF, we compute similarity as specified in \cite{MANTELLI2019304} and label this similarity computation method \emph{mantelli}.

To provide a reference to a method that operates on semantic maps and allows very compact map representation over large areas, we implement the map matching solution proposed in Choi \etal{} \cite{9341682}. We trained a U-Net~\cite{unet} network on the Massachusetts Buildings Dataset~\cite{MnihThesis} to segment buildings in the input images. We employ this network to extract the invariant feature descriptors introduced by Choi \etal{} \cite{9341682} on satellite and \ac{uav}s images: in this setting the feature vectors $\descriptorvector_k$ are built upon building ratio information.

\subsection{Localization performance}\label{ssec:evaluating_localization_performance}
\subsubsection{Experimental setting}
We perform all localization experiments on datasets that contain images collected with a \ac{uav}. In all experiments, we update our belief after at least $u_{l} = $ \translationbetweenupdates{} m of travel have occurred since the previous update. In all experiments, we use a map grid with resolution $r_{xy}=10$ m, $r_{\theta}=6 \degree$.

\subsubsection{\ac{ahrs} and \ac{vio} measurements}
The datasets used in experiments do not contain \ac{imu} or magnetometer measurements and thus we have to simulate them. We simulate odometry measurements by random sampling from distribution \eqref{eq:odometry_posterior_density}. The translation standard deviation $\sigma_{u,xy}$ is approximated as 0.05 m per meter of travel and heading standard deviation $\sigma_{u,\theta}$ as $0.15 \degree$ per meter of travel, which approximately correspond with the performance of \ac{vio} algorithms reported in literature \cite{8460664}.

For heading, we sample from the distribution specified in \eqref{eq:heading_measurement}.  Manufacturers of compact commercial \ac{ahrs} sensors typically report \ac{rms} error of $2\degree$ in heading \cite{xsens_mti3_datasheet, vectornav_vn_100_datasheet}. In all experiments with \ac{ahrs}, we simulate the heading measurement from \ground{} orientation data and assume $\sigma_{v} = 3\degree$.

\subsubsection{Evaluating localization performance}
\begin{figure*}[t!]
    \centering
    \begin{subfigure}[t]{0.5\textwidth}
        \centering
        \includegraphics[width=\linewidth,valign=t]{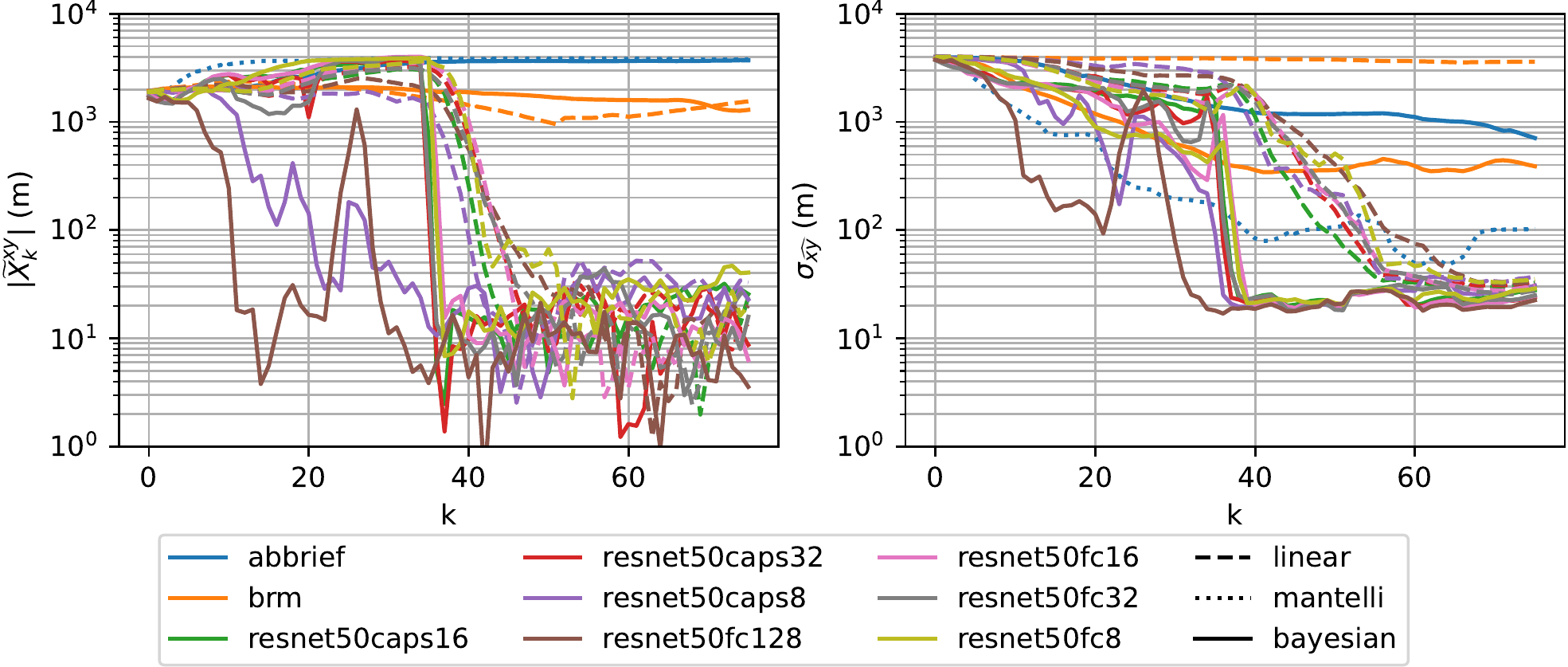} 
        \caption{Dataset 1}
        \label{fig:dataset1_graph}
    \end{subfigure}%
    ~ 
    \begin{subfigure}[t]{0.5\textwidth}
        \centering
        \includegraphics[width=\linewidth,valign=t]{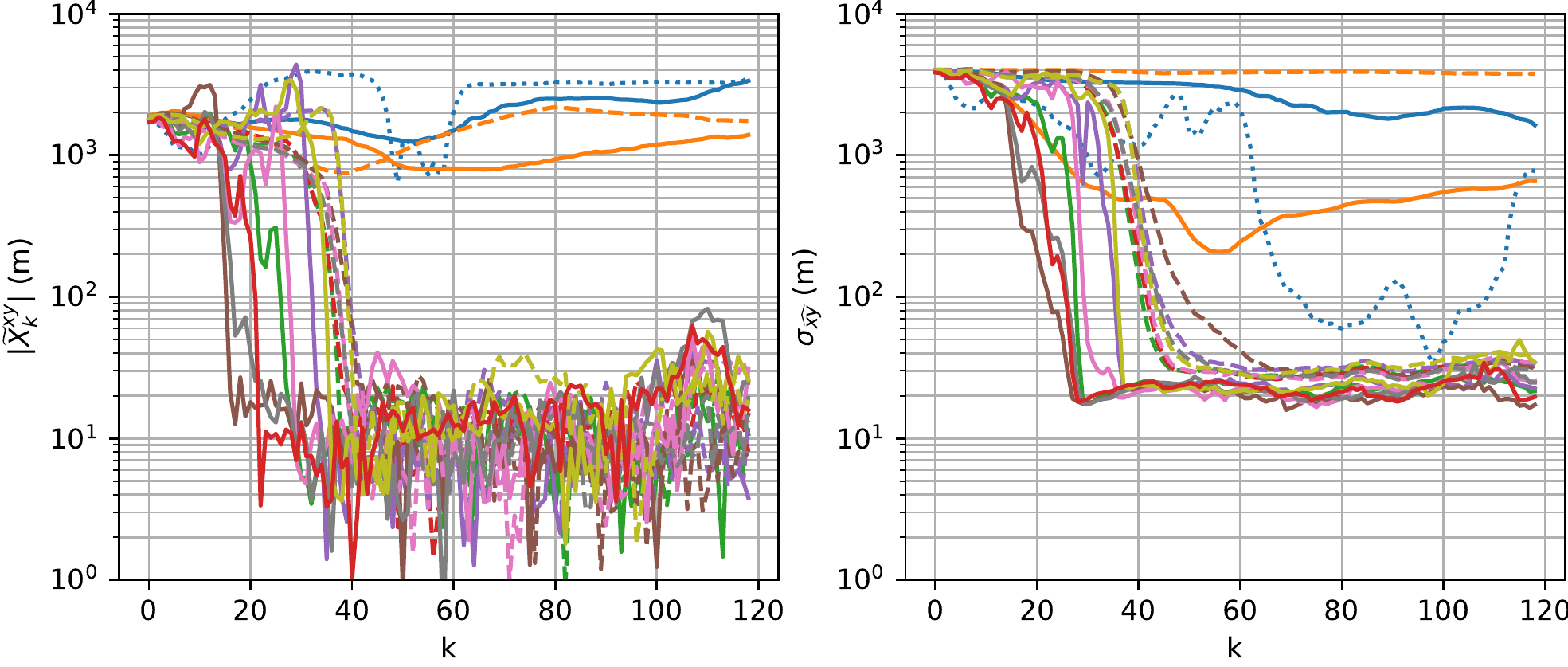} 
        \caption{Dataset 2}
        \label{fig:dataset2_graph}
    \end{subfigure}%
    \hfill
    \begin{subfigure}[t]{0.5\textwidth}
        \centering
        \includegraphics[width=\linewidth,valign=t]{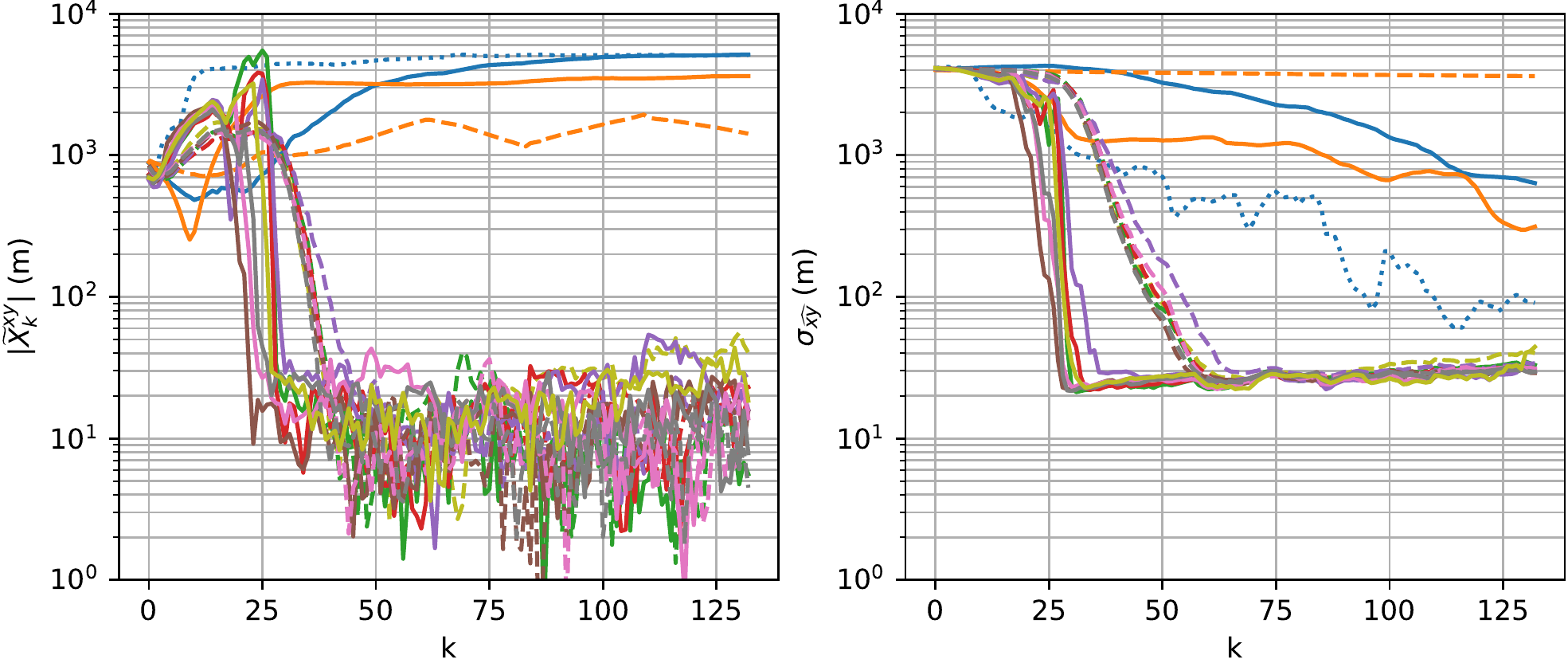} 
        \caption{Dataset 3}
        \label{fig:dataset3_graph}
    \end{subfigure}%
    ~
    \begin{subfigure}[t]{0.5\textwidth}
        \centering
        \includegraphics[width=\linewidth,valign=t]{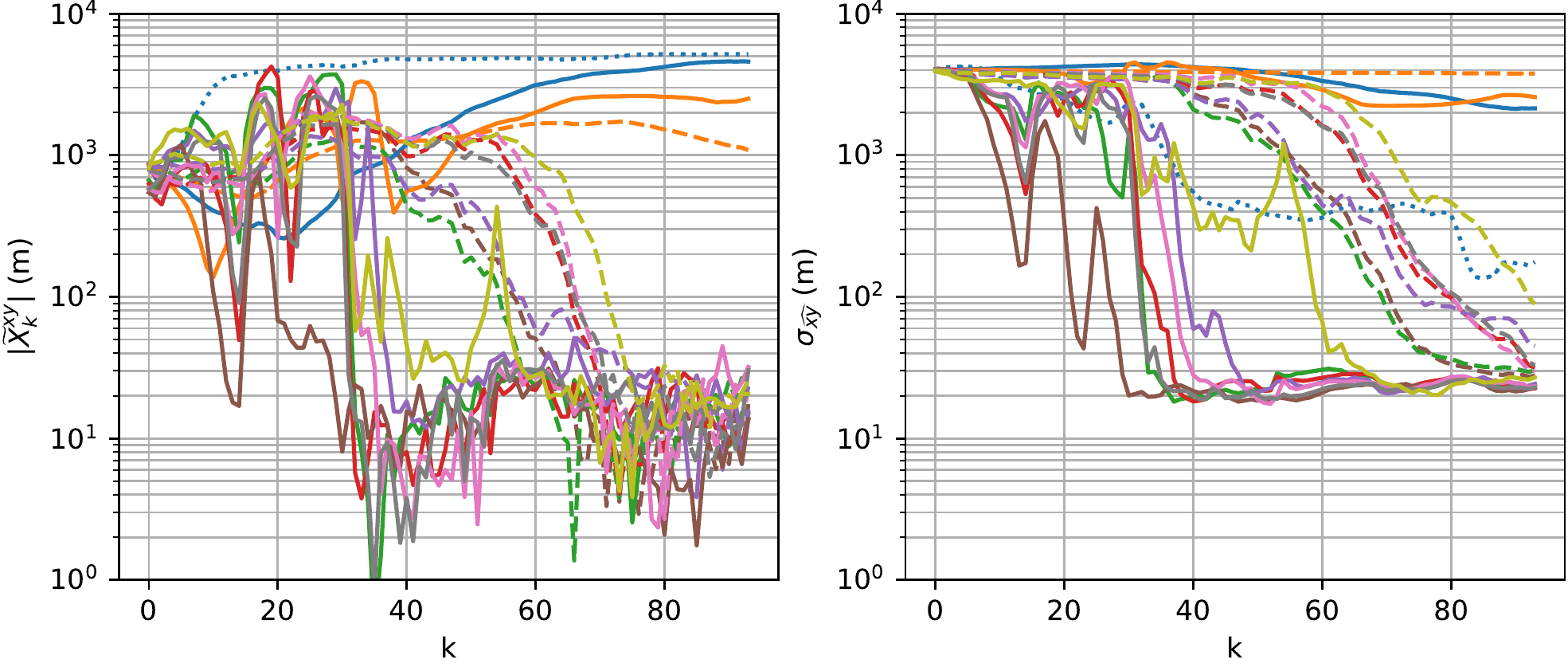} 
        \caption{Dataset 4}
        \label{fig:dataset4_graph}
    \end{subfigure}
    \hfill
    \centering
    \begin{subfigure}[t]{0.5\textwidth}
        \centering
        \includegraphics[width=\linewidth,valign=t]{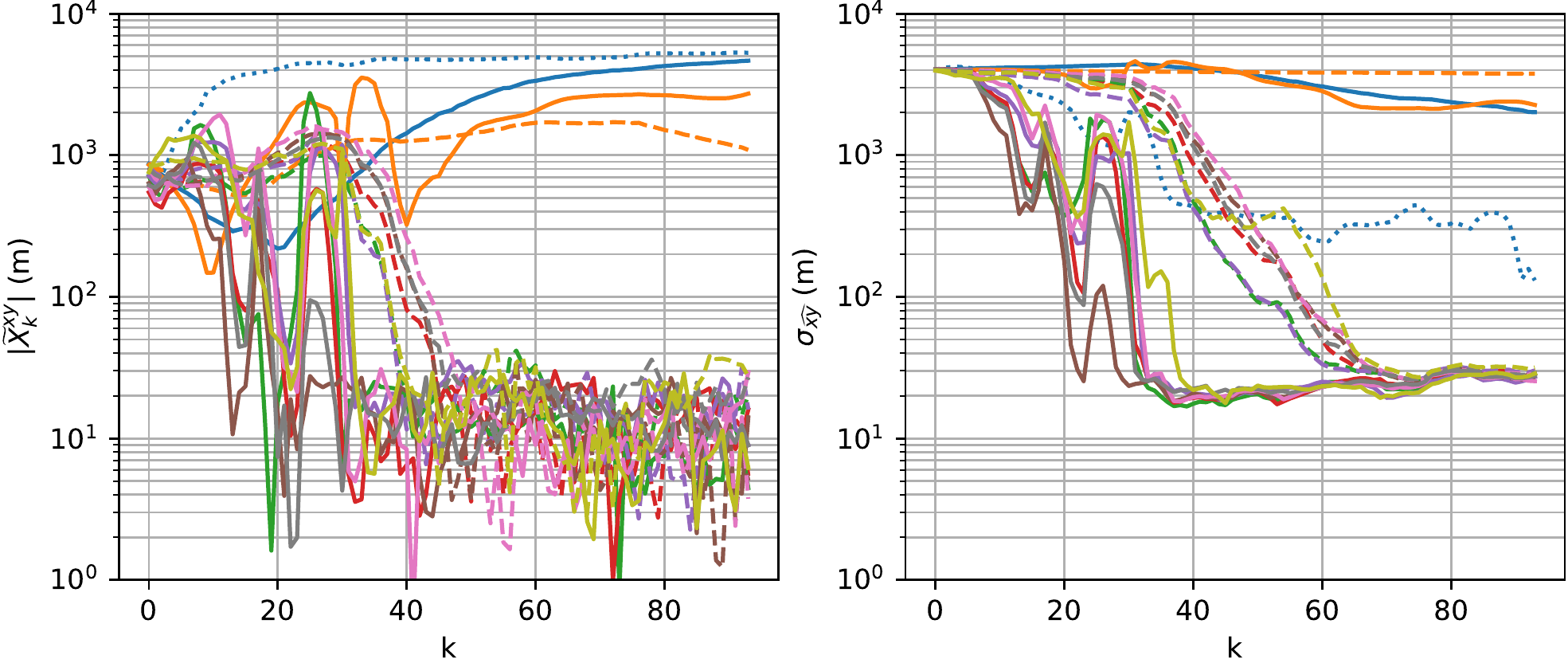} 
        \caption{Dataset 5}
        \label{fig:dataset5_graph}
    \end{subfigure}%
    ~ 
    \begin{subfigure}[t]{0.5\textwidth}
        \centering
        \includegraphics[width=\linewidth,valign=t]{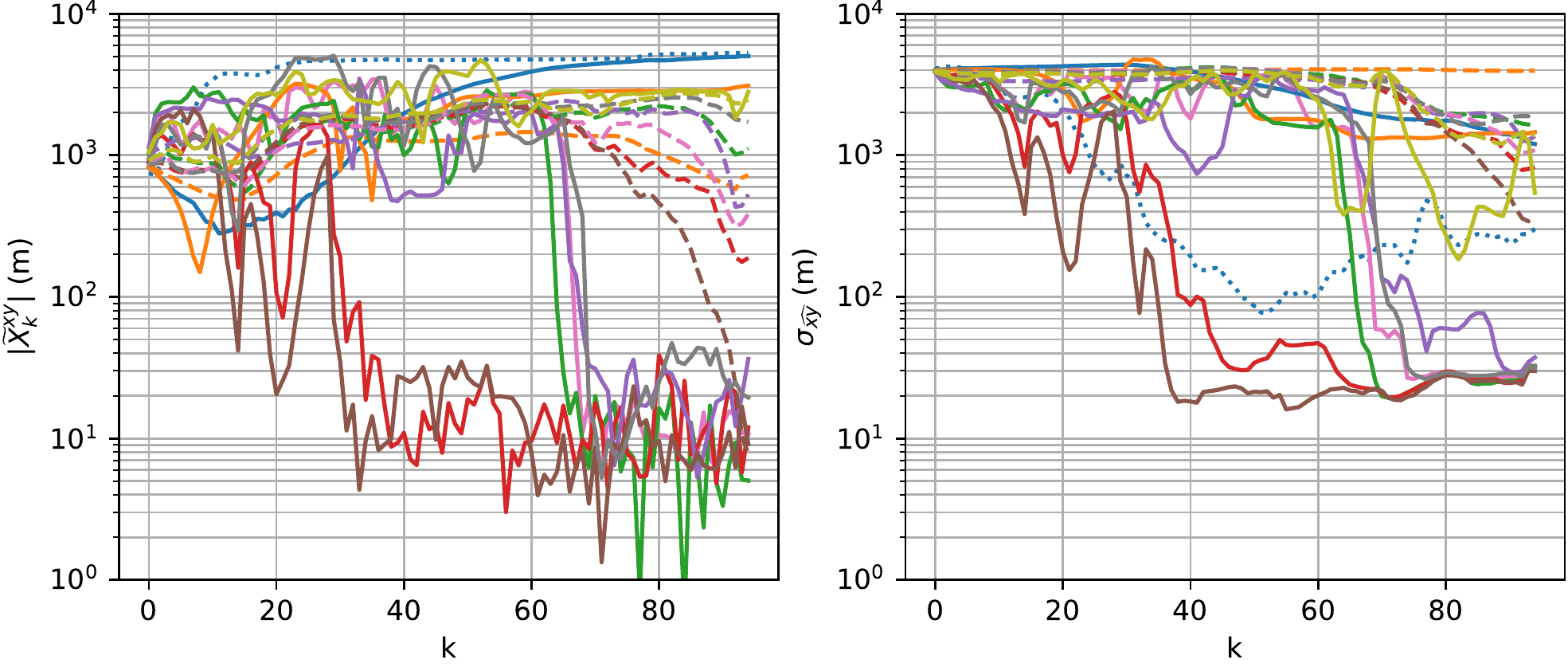} 
        \caption{Dataset 6}
        \label{fig:dataset6_graph}
    \end{subfigure}%
    \hfill
    \begin{subfigure}[t]{0.5\textwidth}
        \centering
        \includegraphics[width=\linewidth,valign=t]{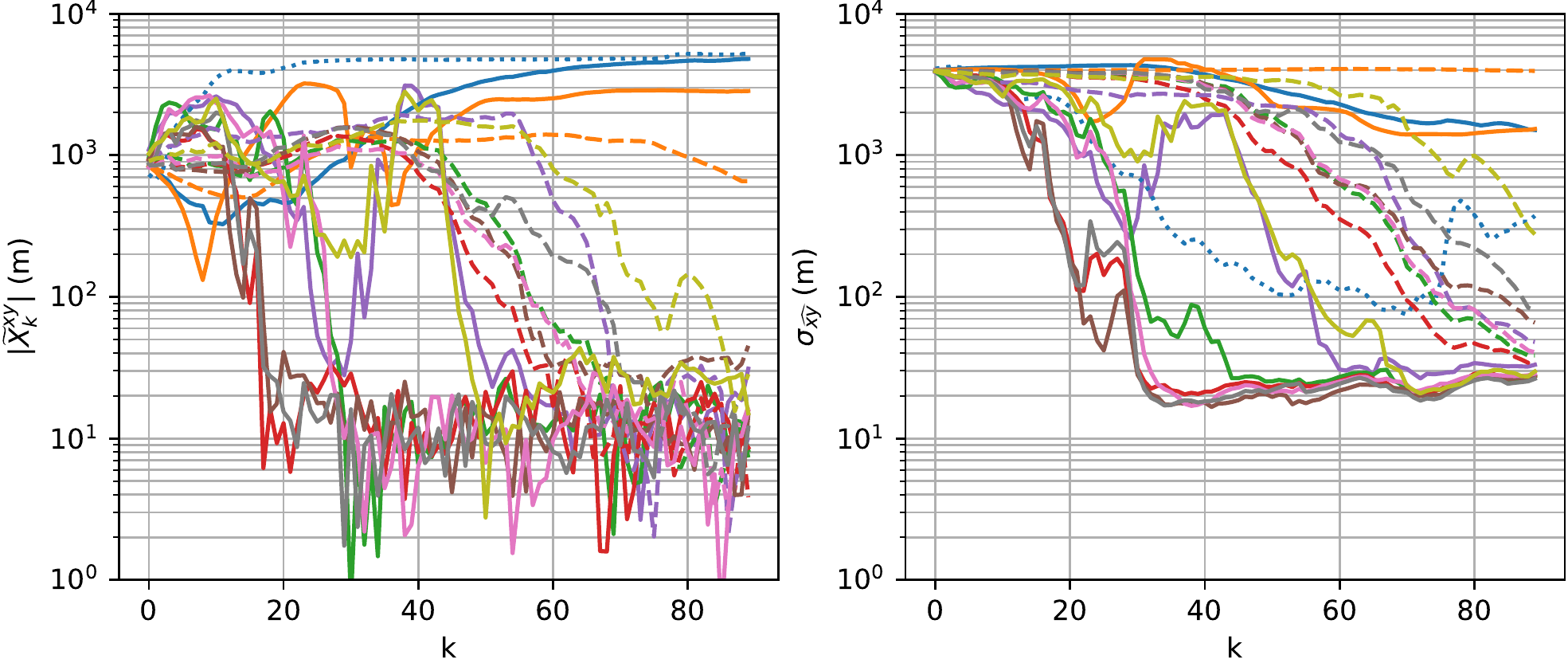} 
        \caption{Dataset 7}
        \label{fig:dataset7_graph}
    \end{subfigure}%
    ~
    \begin{subfigure}[t]{0.5\textwidth}
        \centering
        \includegraphics[width=\linewidth,valign=t]{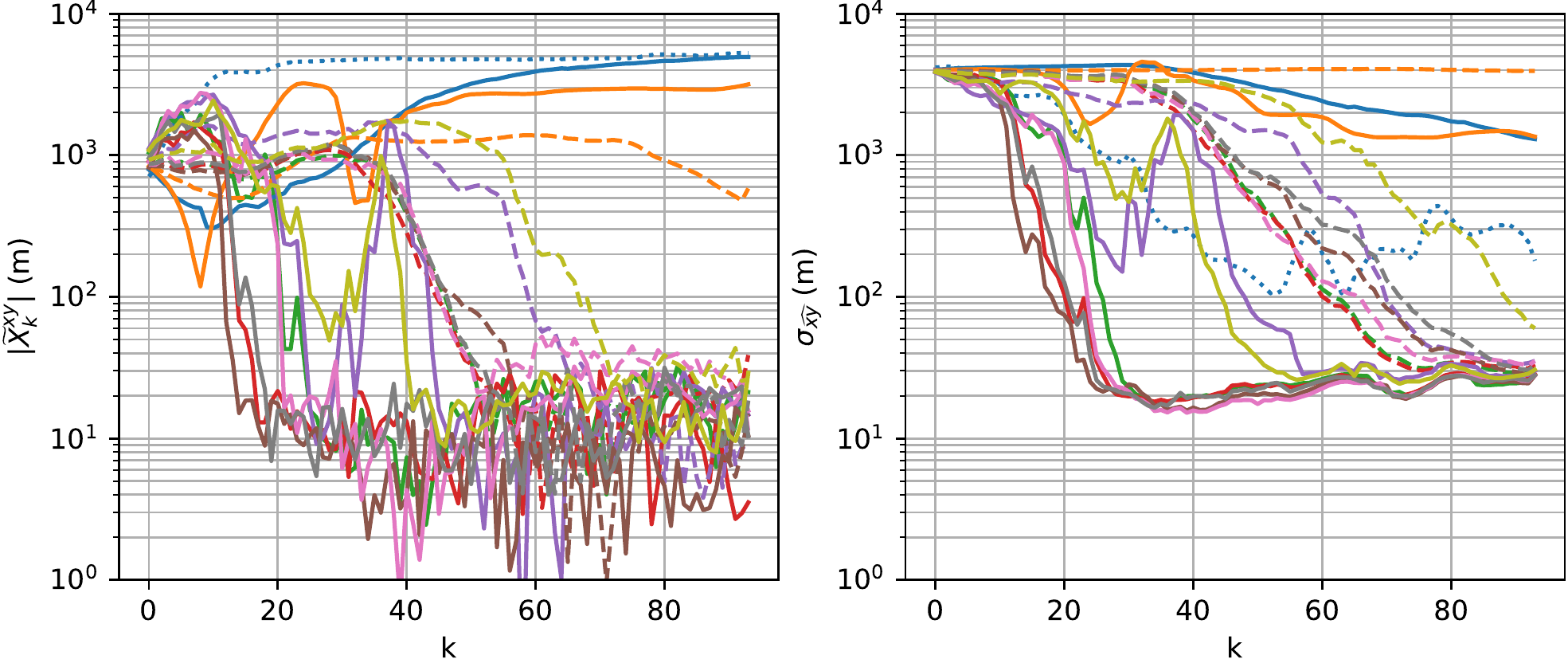} 
        \caption{Dataset 8}
        \label{fig:dataset8_graph}
    \end{subfigure}
    \caption{Translation error $|\widetilde{X}^{xy}_k|$ and standard deviation of translation $\sigma_{\widehat{xy}}$ as function of update index $k$ in different datasets using all methods. Line color represents descriptor computation method, line style (dashed, dotted, solid) represents likelihood computation method. Logarithmic scale. Updates are made approximately every \translationbetweenupdates{} meters of travel.}
    \label{fig:errorplots}
\end{figure*}

We evaluate localization performance for each step $k$ after completing prediction, map matching and \ac{ahrs} update at that step. We define that a localization solution has converged when the translation standard deviation $\sigma_{\widehat{xy}}$ is less than 100 m. We compute the mean number of updates to convergence $\bar{k}_{c}$ and mean translation error in converged state, $\bar{\widetilde{X}}^{xy}_{c}$, for each tested model and likelihood conversion method.  We compute the proportion of flights where each compared solution converges, $p_c$, and tabulate results in \autoref{tab:time_to_convergence_and_error}. In addition, we visualize the translation error and standard deviation of \xy{} translation in each case in \figref{fig:errorplots}.

\begin{table}[]
\caption{Probability of convergence $p_c$, time to convergence $\bar{k}_{c}$ and mean localization error after convergence $\bar{\widetilde{X}}^{xy}_{c}$ when using our methods with various design choices and when comparing to reference method.}
    \centering
    \begin{tabular}{p{0.25\linewidth} p{0.15\linewidth} p{0.1\linewidth} p{0.1\linewidth} p{0.1\linewidth}}
\toprule
\modeltypetableheader & \likelihoodconversiontableheader & $p_{c}$ & $\bar{k}_{c}$ & $\bar{\widetilde{X}}^{xy}_{c} (m)$ \\
\midrule
Ours, Caps, D=8 & Linear & 0.875 & 60.9 & 15.7 \\ 
Ours, Caps, D=16 & Linear & 0.875 & 56.3 & 13.3 \\ 
Ours, Caps, D=32 & Linear & 0.875 & 58.9 & 13.9 \\ 
Ours, FCN, D=8 & Linear & 0.75 & 65.2 & 21.1 \\ 
Ours, FCN, D=16 & Linear & 0.875 & 61.1 & 13.9 \\ 
Ours, FCN, D=32 & Linear & 0.875 & 63.1 & 12.9 \\ 
Ours, FCN, D=128 & Linear & 0.875 & 62.3 & 11.2 \\ 
Ours, Caps, D=8 & Bayesian & 1.0 & 44.4 & 18.3 \\ 
Ours, Caps, D=16 & Bayesian & 1.0 & 35.2 & 12.8 \\ 
Ours, Caps, D=32 & Bayesian & 1.0 & 30.8 & 15.0 \\ 
Ours, FCN, D=8 & Bayesian & 0.875 & 43.0 & 18.7 \\ 
Ours, FCN, D=16 & Bayesian & 1.0 & 36.1 & 14.6 \\ 
Ours, FCN, D=32 & Bayesian & 1.0 & 33.2 & 15.7 \\ 
Ours, FCN, D=128 & Bayesian & 1.0 & 23.2 & 12.6 \\ 
BRM & Linear & 0.0 & N/A & N/A \\ 
BRM & Bayesian & 0.0 & N/A & N/A \\ 
abBRIEF & Mantelli & 0.625 & 63.8 & 4112.2 \\ 
abBRIEF & Bayesian & 0.0 & N/A & N/A \\ 

\bottomrule
    \end{tabular}
    \label{tab:time_to_convergence_and_error}
\end{table}

\subsubsection{Statistical significance testing}
To explore what significance our results show in the various parameters we have used in the experiment configurations, we run a type II analysis of variance test on time to convergence and on localization error after convergence. We consider the embedding dimension (8, 16, 32, 128), projection module type (fully connected or capsules) and likelihood computation method (linear or bayesian) and combinations of these parameters. We use $p=0.05$ as limit for significance.

For time to convergence, we reject null hypothesis that chosen likelihood method is not significant ($p=4.2*10^{-16}$). For localization error after convergence, we reject null hypothesis that the parameter is not significant for
embedding dimension ($p=2.5*10^{-14}$),
projection module type ($p=2.1*10^{-2}$),
likelihood method ($p=3.7*10^{-4}$)
and combination of embedding dimension and projection module type ($p=5.0*10^{-5}$) and combination of embedding dimension and likelihood method ($p=1.3*10^{-2}$).

\subsection{Real-time experiment}
\label{sec:realtime_experiment}

\begin{figure}
\centering
\includegraphics[width=0.7\linewidth]{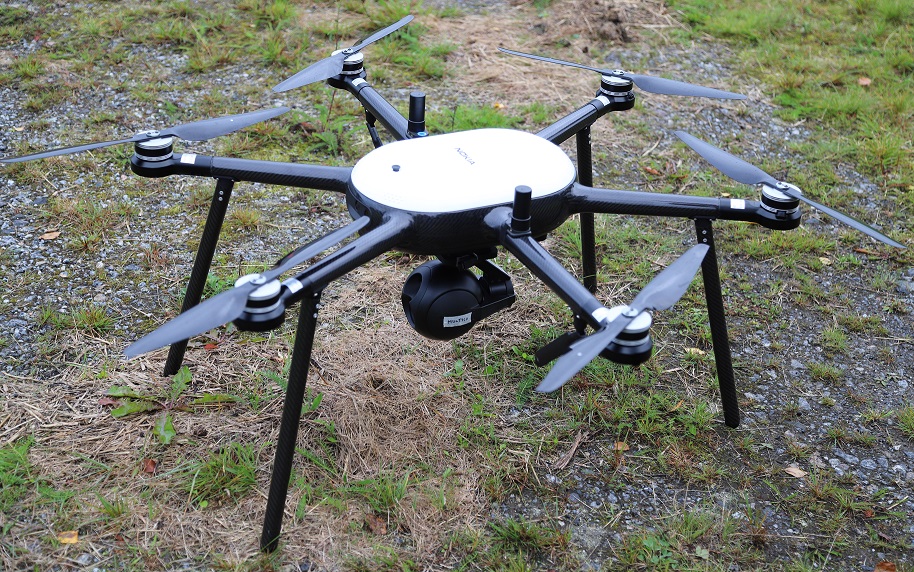}
\caption{\ac{uav} used in real time experiment}
\label{fig:image_of_nokia_drone}
\end{figure}

To understand the applicability of our solution to an embedded system, we implemented a version of our algorithm running in real time on an embedded computer on an example \ac{uav}. We use the Nokia Drone Networks drone (see \autoref{fig:image_of_nokia_drone}) carrying a camera gimbal. The drone was equipped with an NVidia Jetson Nano computer on which the localization algorithm was run. The algorithm was implemented in Python and we used ROS \cite{ros} for inter-process communication. To accommodate the smaller observable ground footprint due to the narrower field of view of the drone camera at the selected flight altitude and observation parameters (50 m altitude at 45 degree camera pitch), we retrained a model with Resnet50 + FCN, D=16 architecture to work on 40m by 40m images at 1 m/px resolution. The size of the operating region and map was \kauhavamapdimensions{}. We used VINS-Mono \cite{8421746} as the \ac{vio} algorithm. We used the output of a non-GNSS-corrected \ac{ahrs} algorithm implemented on the drone flight controller for heading updates.

The mean network inference time (run on CPU) was 1.02 s, mean prediction time was 0.83 s, map matching took 1.18 s and \ac{ahrs} update took on average 0.12 s, and all steps took on average 3.15 s. The algorithm was configured to perform an update every 40 meters of travel, while the drone was flying at 5 m/s. There is room for speedup by proper parallelization of the algorithm. The algorithm was run fully onboard the \ac{uav}.

Some localization errors appeared over terrain patches with very uniform colouring due to scale drift and rotation estimation errors of the \ac{vio} algorithm which are not considered in our odometry noise model. In addition, the \ac{ahrs} heading estimate from the flight controller appeared to contain non-Gaussian heading errors up to 15 degrees; to accommodate this, we used $\sigma_{v} = 60\degree$ in this experiment.

The memory footprint of the precomputed map in this experiment was 460.8 MB. On a \maparea{} map, the precomputed map memory requirement for D=8, D=16, D=32 and D=128 are 3.5 GB, 7.0 GB, 14.0 GB and 55.9 GB, respectively, setting further practical constraints for embedded implementation at very large scale. 

This experiment shows that the algorithm can be run in real time on an embedded computer carried by a drone in an operating region of typical size for drone operations.

\section{Discussion}
\label{sec:discussion}

To solve the wake-up robot problem at the presented scale, we derived a solution based on a recursive point mass filter instead of the more common particle filter approach. We believe this avoids issues with particle depletion, which is important in cases of significant initial uncertainty, ambiguity of terrain and potential intermittent but not random correspondence mismatches between observations and maps.

An architecture where an image observation is projected into an embedding space, with a major reduction in data dimensionality, is a key enabler for fast and memory-efficient similarity comparisons over a large number of hypotheses. Our approach for projecting data from different source domains (\ie{} \ac{uav} camera images and orthophoto maps or satellite images) into a common domain enables the use of learned descriptors. With our approach, extensive training data covering all expected variation is required in only one source domain without need for labeling of data. Our approach of using a learnable embedding appears to be efficient for localization over large areas containing natural and built environments, in scenarios where also significant seasonal appearance change occurs between flights. Other existing methods, \eg{} a handcrafted descriptor approach (abBRIEF) or learned description trained to detect pre-specified semantics (\ac{brm}), do not converge at this scale.

Bayesian likelihood conversion showed the greatest effect in time to convergence, in comparison to the more common linear approximation. The results in \tabref{tab:time_to_convergence_and_error} appear to hint at the possibility that increasing embedding dimensionality leads to faster convergence, but further experimentation would be required to verify statistical significance. For localization error afer convergence, embedding dimensionality was found to be one of the statistically significant parameters. Results tabulated on \tabref{tab:time_to_convergence_and_error} and error plots in \figref{fig:errorplots} seem to suggest that there may be a lower bound on localization error that appears independent from dimensionality and is most likely a result of other design choices and the characteristics of the operating environment. In other words, it appears that low-dimensionality descriptors are an efficient way of expressing what is important for localization and dimensionality of description is not the main hindering factor, what comes to localization error.

In order to extract low-dimensionality descriptors, we tested both traditional fully connected and capsule layers.
Our hypothesis was initially that capsule layers would improve localization performance, extracting better encodings thanks to their ability to model part-whole relationships and robustness to novel viewpoints.
Employing capsule layers lead to a slight improvement in error after convergence in comparison to fully connected layers while using less trainable parameters, as also stated in~\cite{hinton-dynamic}. In fact, for 16-dimensional embeddings, with fully connected layer, the network has 57M trainable parameters, while with capsule layers, it has ~42M trainable parameters. Since this improvement is only marginal, we did not conduct experiments with 128-dimensional capsule embeddings, as CapsNets are also well known to be  computationally very demanding in terms of memory consumption, training and inference times.

Experiments demonstrate that the proposed approach is suitable for real time implementation on a flying platform at a typical scale of \ac{uav} operations. Tailoring design parameters allows the implementation of the solution on a very resource-constrained platform and running it in real time together with an odometry system.

To enable error recovery, the solution provides a measure of position uncertainty by computation of standard deviation. Upon detection of prolonged high standard deviation, this enables the triggering of reinitialization of the pose estimator. The ability to detect localization failures and recover from the loss of location information on a large scale paves way for a failure-aware, failsafe \ac{uav} localization system.

\section{Conclusions}
\label{sec:conclusions}

We have shown that the approach utilizing our map matching method together with a point mass filter is able to resolve \ac{uav} pose even in the case of highly uninformed initialization corresponding to a map size of \maparea{}, in conditions of significant seasonal appearance change between \ac{uav} image and map, even when flying over areas with natural ambiguity. The proposed solution converges to a localization error of \translerrorafterconvergence{} on average in \numberofstepstoconvergence{} updates, depending on the chosen architecture, while reference methods are not able to converge to the correct pose under the same circumstances. All of these contributions show that real-time localization is possible on a large scale. Going beyond the demonstrated \maparea{} will require being able to represent even larger hypothesis spaces. 
Addressing this challenge will potentially require future work in hierarchical models in order to retain the favorable characteristics of high spatial accuracy, extreme initial uncertainty, and complete coverage of the hypothesis space.

\section*{Acknowledgment}

We thank Olli Knuuttila for the implementation work in real-time experiments. We thank Mika J\"{a}rvenp\"{a}\"{a} and Kari Hautio for testing opportunities and support with implementation on the Nokia Drone Networks device.

\bibliographystyle{./bibliography/IEEEtran}
\bibliography{./bibliography/IEEEabrv,./bibliography/bibliography}

\end{document}